%% file: template.tex
\documentclass{article}

\usepackage{arxiv}

\usepackage[utf8]{inputenc} 
\usepackage[T1]{fontenc}    
\usepackage{hyperref}       
\usepackage{url}            
\usepackage{booktabs}       
\usepackage{amsfonts}       
\usepackage{nicefrac}       
\usepackage{microtype}      
\usepackage{lipsum}
\usepackage{graphicx}
\usepackage{hyperref}
\usepackage{times}
\usepackage{epsfig}
\usepackage{amsmath}
\usepackage{chngcntr}
\usepackage{cellspace, makecell, multirow}
\usepackage{textcomp}
\usepackage{longtable}
\usepackage{xspace}
\usepackage{bookmark}
\usepackage{subcaption}
\usepackage{algorithm}
\usepackage{algpseudocode}
\usepackage{afterpage}
\usepackage{tikz}
 \usepackage{placeins}

\title{Impact of imperfect annotations on CNN training and performance for instance segmentation and classification in digital pathology}

\title {Impact of imperfect annotations on CNN training and performance for instance segmentation and classification in digital pathology}

\author{Laura G\'alvez Jim\'enez \thanks{Corresponding author.}\\
	LISA\\
	 Universit\'e Libre de Bruxelles\\
	Belgium \\
	\texttt{laura.galvez.jimenez@ulb.be} \\
	\and
	 Christine Decaestecker\\
      LISA and CMMI\\
	 Universit\'e Libre de Bruxelles\\
	Belgium  \\
	\texttt{christine.decaestecker@ulb.be} \\
}

\begin{document}
\maketitle
\begin{abstract}
Segmentation and classification of large numbers of instances, such as cell nuclei, are crucial tasks in digital pathology for accurate diagnosis. However, the availability of high-quality datasets for deep learning methods is often limited due to the complexity of the annotation process. In this work, we investigate the impact of noisy annotations on the training and performance of a state-of-the-art CNN model for the combined task of detecting, segmenting and classifying nuclei in histopathology images. 
In this context, we investigate the conditions for determining an appropriate number of training epochs to prevent overfitting to annotation noise during training. Our results indicate that the utilisation of a small, correctly annotated validation set is instrumental in avoiding overfitting and maintaining model performance to a large extent. Additionally, our findings underscore the beneficial role of pre-training.
\end{abstract}


\input{introduction.tex}

\input{relatedworks.tex}

\input{materialsmethods.tex}

\input{results.tex}

\input{conclusions.tex}

\newcommand{\printcredits}

\bibliographystyle{unsrt}  
\bibliography{cas-refs}  



\input{appendix}
\end{document}

%% file: introduction.tex
\section{Introduction}

One of the main challenges encountered in machine and deep learning applied to image segmentation and classification concerns the presence of noisy or imperfect annotations in datasets, which is why many studies focus on techniques to address this problem \cite{frenay2013classification,nigam2020impact,algan2021image}. The annotation noise takes different forms depending on the task and the supervision errors it can generate. In a segmentation task for instance, these errors relate to ill-defined contours or missing objects \cite{8759545}, whereas in image classification, the errors typically involve class labels \cite{xiao2015learning,saez2016influence}. 

Medical applications can be heavily impacted by supervision errors due to the intricate nature of the annotation process, which is not as straightforward as in other types of applications. In digital pathology, whole slide images are of considerable size, with a large number of instances to annotate.
This complexity and the time-consuming nature of the annotating process are exacerbated by the fact that not all annotators are necessarily experts; they may even be students supervised by a single expert \cite{verma2021monusac2020}, thereby reducing the reliability of the annotation process.
Furthermore, the common occurrence of inter- and intra-observer variability in the clinical field adds another layer of complexity, making it challenging to obtain trustworthy annotations.
Although strategies for generating a consensus exist \cite{warfieldStaple}, the absence of a definitive ground truth, unlike in other fields, poses an additional challenge. 
To address this issue in medical imaging, numerous investigations have been conducted within the context of image segmentation or classification \cite{karimi2020deep}.

As detailed in the next section, no study has addressed the challenge of handling imperfect annotations in the context of detecting, segmenting, and classifying multiple instances in medical images. The present study aims to fill this gap by tackling this issue as a classical digital pathology problem, specifically the detection, segmentation, and classification of different types of cell nuclei in tissue samples of various origins.
Our methodology involves using training sets in which we intentionally introduce annotation errors, affecting each of these tasks in a controlled manner. Within this framework, the objectives are to study the impact of these annotation errors on the performance of a state-of-the-art CNN model, and to examine the conditions necessary to avoid overfitting to annotation noise during its training.

The main contributions of this study are as follows: Our analysis of the impact of imperfect annotations on the training process highlights the challenges in determining the appropriate number of epochs before overfitting to annotation noise, especially in the absence of a clean (i.e., correctly annotated) validation set. Additionally, we outline the negative impact of this overfitting on the performance, assessed by specific metrics for each of the tasks considered (detection, segmentation, and classification). Finally, our experiments demonstrate that significant performance recovery is possible by stopping training based on model evaluation using a small but clean validation set. They also underscore the beneficial role that pre-training can play.

This article is organised as follows: In Section \ref{sec:related_works}, we briefly review related works. 
In Section \ref{sec:material_methods}, we describe the datasets used in our experiments, the methodologies for introducing imperfect annotations, the model used, as well as the training algorithm, including the training stopping strategy, and finally, the performance evaluation metrics. In Section \ref{sec:results}, we describe the experiments and present the results. 
We conclude with a discussion and final remarks in Sections \ref{sec:discussion} and \ref{sec:conclusion}.

%% file: relatedworks.tex
\section{Related works}
\label{sec:related_works}
Imperfect or erroneous annotations are observed in several fields, each with its own specificity. In this section, we concentrate on the digital pathology domain as much as possible. We review works that study the nature of imperfect annotations and their impact on model performance. We also explore studies that investigate methods to avoid overfitting to annotation noise during deep network training.

\subsection{Impact of imperfect annotations} 
\label{section:review_imperfect}

Karimi et al. \cite{karimi2020deep} identified three main sources of annotation noise in medical images: inter-observer variability, annotator error, and errors in computer-generated labels. In the context of medical image segmentation, Tajbakhsh et al. \cite{tajbakhsh2020embracing} categorised imperfect annotations as either scarce or weak, while Frénay and Verleysen \cite{frenay2013classification} developed a taxonomy of label noise in classification, distinguishing between completely random noise, random noise, and non-random noise. 

To investigate the impact of imperfect annotations on histopathology image segmentation, Foucart et al. \cite{8759545} introduced the concept of ``SNOW'' dataset, representing Semi-supervised, Noisy, and/or Weakly annotated datasets. These authors measured the impact of imperfections by simulating missing and imprecise annotations on gland segmentation performance. Similarly, V{\u{a}}dineanu et al. \cite{vuadineanu2022analysis} studied the impact of imperfect annotations on cell segmentation, employing a synthetic dataset and emulating imperfections by removing and adding cells or perturbing their borders. 
Both studies agree on the robustness of CNNs up to a certain level of missing annotations, but observe a clear drop in performance when the percentage of omissions becomes high. Additionally, biased or fairly imprecise contours, such as bounding boxes, also impact performance. Finally, the impact of erroneous inclusions in a synthetic dataset demonstrates dependency on object size \cite{vuadineanu2022analysis}.

More complex tasks that involve the combination of detection, segmentation, and classification of instances, such as cell nuclei or glands, are also of interest in digital pathology \cite{nir2018automatic,verma2021monusac2020,barmpoutis2022digital}. In this context, errors or imperfections in annotations can manifest in different forms, affecting both the image background (such as missing annotations) and foreground (such as instance class errors). While annotation noise has been extensively studied in the context of image classification \cite{algan2021image}, it has not been thoroughly explored for the classification of multiple segmented instances within images. Yang et al. \cite {yang2020learning} investigated instance segmentation and classification in the presence of noisy class labels, but in a video application context, introducing features not commonly encountered in medical imaging. 

To the best of our knowledge, the impact of various types of annotation noise on the combined task of detecting, segmenting and classifying multiple instances within images has not been investigated, especially in the context of medical imaging and digital pathology. This gap motivated the present study.

\subsection{Approaches to avoid overfitting to annotation noise during training}
\label{section:review_approaches}
As reported below, almost all the work in the literature focuses on segmentation or image classification tasks, with very few attempting to address this problem in the context of multi-class instance segmentation and classification.

To the best of our knowledge, the need for early stopping of deep network training in the presence of incorrect supervision has only been analysed for image classification tasks outside the medical domain. For instance, Song et al. \cite{song25does} studied the impact of noisy labels in the CIFAR dataset on the network training process and emphasised the need to halt training before the network becomes overly impacted by noise. They also highlighted the importance of having a clean validation set for determining the stopping point. In the same context of label-noise learning, Bai et al. \cite{bai2021understanding} pursued the idea of early training stopping but applied it sequentially to different parts of the network, rather than applying it globally to the whole network and showed the superiority of their approach over other state-of-the-art methods. These findings prompted us to investigate the extent to which such an approach would also provide a relevant solution for a more complex task involving the detection, segmentation, and classification of multiple instances in the presence of different sources of annotation noise.

An alternative is to attempt to identify and remove (or clean) erroneous annotations. Karimi et al. \cite{karimi2020deep} focused on label cleaning for brain lesion segmentation, and Li et al. \cite{li2022study} performed data pruning to clean an esophageal endoscopy dataset. Recently, we demonstrated the capabilities of model prediction-based filtering to remove data with noisy annotations from the original MoNuSAC training set, resulting in a positive impact on model performance \cite{galvez2023cleaning}. In the present study, we make use of the resulting cleaned training set to subsequently introduce annotation errors in a controlled way.

Semi-supervised approaches can also be used, provided that enough correct annotations can be identified with a certain degree of certainty for the supervised stage. However, this condition is not automatically guaranteed, especially in the presence of noisy classification labels. While these approaches may be appropriate for addressing the issue of missing annotations  \cite{lai2021semi,pulido2020semi,wang2020unlabeled}, other studies on segmentation show that they do not necessarily yield the expected results \cite{foucart2020snow,jimenez2023computational}. In particular, the latter studies reveal that these approaches offer no significant improvement over the simple approach of focusing model training on image areas with annotations. Finally, weakly supervised approaches, based on the simple labelling of the presence/absence of objects of interest within thumbnails, are ill-suited when small objects (such as cell nuclei) from different classes can agglomerate in the same image. For these reasons, these approaches are not addressed in the present study, as also discussed in Section \ref{sec:discussion}.

%% file: materialsmethods.tex
\section{Material and methods}
\label{sec:material_methods}
\subsection{Datasets} \label{section:datasets}
\subsubsection{Description}
In this study, we use two datasets, MoNuSAC and PanNuke, which we simplify to some extent, as detailed below.

For the purposes of the present work, we require sufficiently well-annotated datasets in which we can introduce different types of annotation corruptions in a controlled manner (see Section \ref{section:imperfections}). Initially, we employ the clean V1.2 subset extracted from the MoNuSAC training set using a model prediction-based filtering strategy to select more reliable training data \cite{galvez2023cleaning}. As detailed in this latter study, each image in the dataset has a segmentation mask and a classification mask (see Figure \ref{fig:errors_example}a-c). The dataset consists of four cell classes (epithelial, lymphocyte, neutrophil, and macrophage) from four tissue types. Due to the difficulty of segmenting macrophages, we exclude them, considering them as part of the background.
The test set includes an additional class called "ambiguous", which collects image regions containing nuclei with poorly defined boundaries or uncertain class labels to annotators \cite{verma2021monusac2020}.
To maintain reliable annotations for evaluation purposes, we exclude these ambiguous regions from the test set. Validation sets are extracted from the test set, as explained in Section \ref{sec:TrainingStopping}.

In addition, we use the PanNuke dataset \cite{gamper2020pannuke}. It consists of patches extracted from whole-slide images encompassing 19 tissue types, each containing five distinct cell classes (neoplastic, non-neoplastic epithelial, inflammatory, connective, and dead).
Similar to MoNuSAC's macrophages, we exclude dead cells from the annotations, considering them as belonging to the background, due to the huge imbalance between classes, resulting in poor model performance for these cells. The PanNuke dataset is divided into three folds, of which we use two for training and one for validation and testing (see Section \ref{sec:TrainingStopping}). We work with the original PanNuke data as the annotation process appears to be more reliable (a CNN generated the annotations, which were then verified by a panel of experts), and applying the cleaning method developed in \cite{galvez2023cleaning} has proven to be ineffective. 

Table \ref{Tab:MonusacvsPannuke} compares the numbers and types of annotated cell nuclei in both datasets before any preprocessing. It is worth mentioning that the PanNuke classification task presents increased difficulty, as one cell type, such as inflammatory, can cover several sub-types that do not necessarily have the same characteristics.

\begin{table*}[!ht]
    \centering
    \caption{Composition of the V1.2 MoNuSAC and PanNuke sets used.}
    \begin{tabular}{@{}c|cc|cc@{}}
    \multicolumn{1}{c}{} &\multicolumn{2}{c|}{Training} &\multicolumn{2}{c}{Test/Validation} \\ 
    \multicolumn{1}{c|}{Cell type} & 
    \multicolumn{1}{c}{V1.2 MoNuSAC} &
    \multicolumn{1}{c|}{PanNuke} & 
    \multicolumn{1}{c}{V1.2 MoNuSAC} &
    \multicolumn{1}{c}{PanNuke} \\
    \hline
    \multirow[c]{7}{*}{}
    Epithelial & 13558 & 17687 & 7209 & 8867\\
    Lymphocyte & 13389 & - & 7803 & - \\
    Neutrophil & 563 & - & 172 & - \\
    Connective & - & 33702 & - & 16692\\
    Inflammatory & - & 21493 & - & 10571\\
    Neoplastic & - & 54662 & - & 22722\\
    \hline
    Total & 27501 & 127544 & 15184 & 58852\\

    \end{tabular}
\label{Tab:MonusacvsPannuke}
\end{table*}

\subsubsection{Preprocessing: tiling, augmentation and class balancing}
For MoNuSAC, we apply tiling, data augmentation, and class balancing procedures as detailed in our previous work \cite{galvez2023cleaning}. In brief, we extract 256x256 patches with overlap for training and without it for testing and validation. The PanNuke dataset comprises 256x256 non-overlapping patches used as is. For training data augmentation in both datasets, we apply random rotations, flips, hue, saturation, contrast, and brightness perturbations. This process aims not only to increase the amount of training data but also to enhance the network's robustness to potential variations in tissue staining and image acquisition. Additionally, we ensure that each class is represented in all training batches using a weighted sampler, as detailed previously \cite{galvez2023cleaning}.

\subsection{Annotation corruption in clean training sets} \label{section:imperfections}
\begin{figure*}[!ht]
            \centering
            \begin{subfigure}[t]{0.3\textwidth}
            \includegraphics[width=\linewidth]{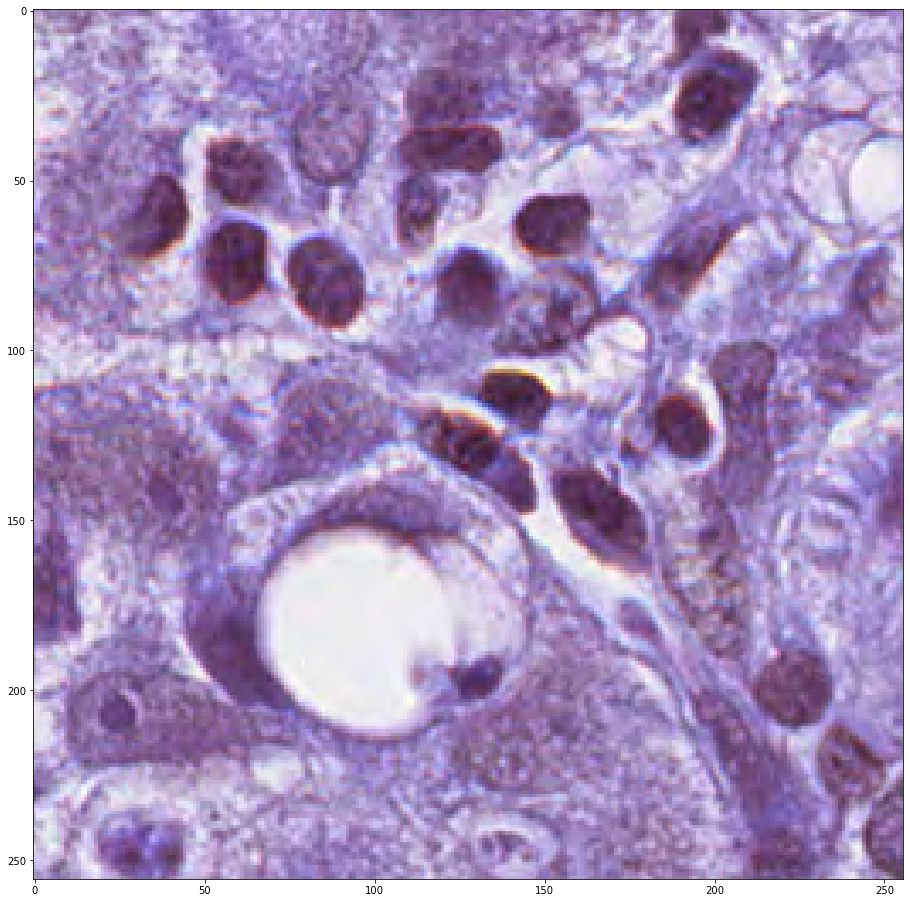}
            \caption{Original patch}
            \label{fig:original_image}
            \end{subfigure}
            \begin{subfigure}[t]{0.3\textwidth}
            \includegraphics[width=\linewidth]{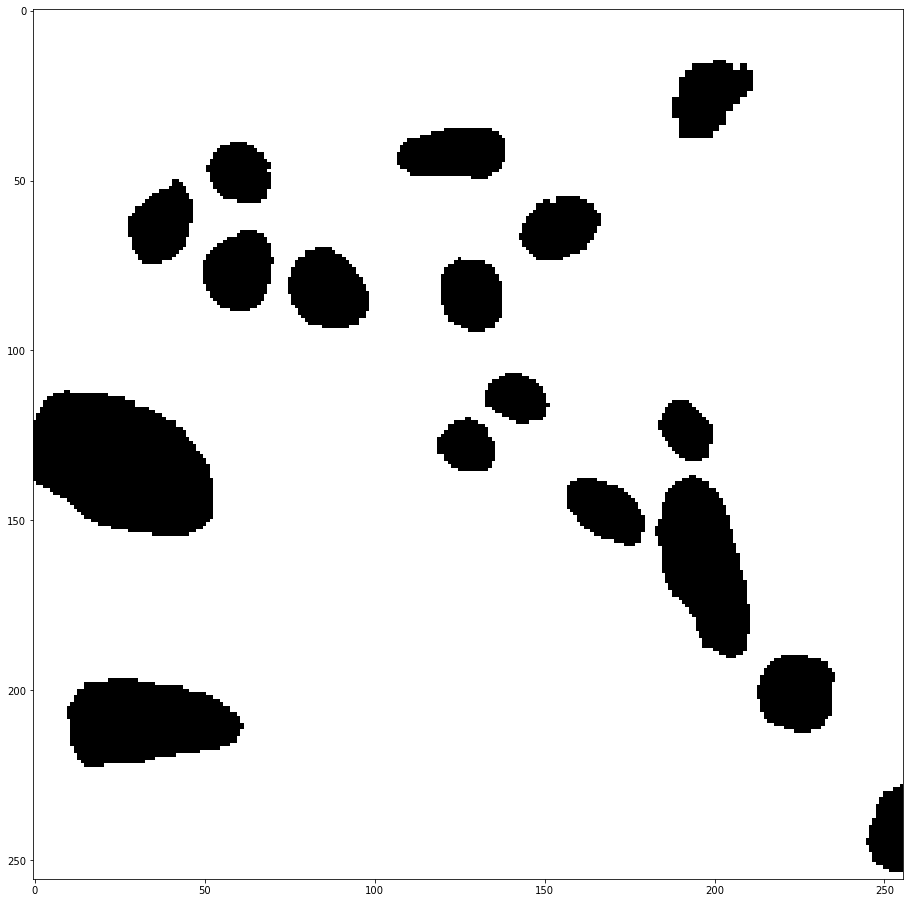}
            \caption{Original segmentation mask}
            \label{fig:original_seg_mask}
            \end{subfigure}
            \begin{subfigure}[t]{0.3\textwidth}
            \includegraphics[width=\linewidth]{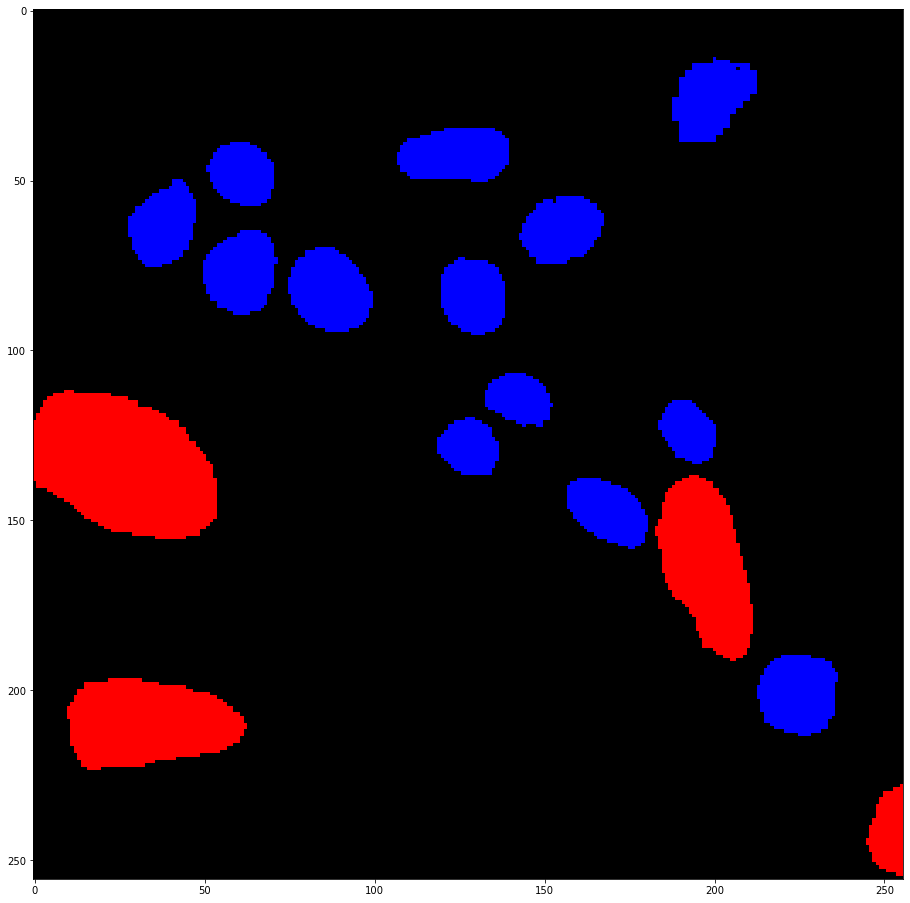}
            \caption{Original classification mask}
            \label{fig:original_class_mask}
            \end{subfigure}
            \begin{subfigure}[t]{0.3\textwidth}
            \includegraphics[width=\linewidth]{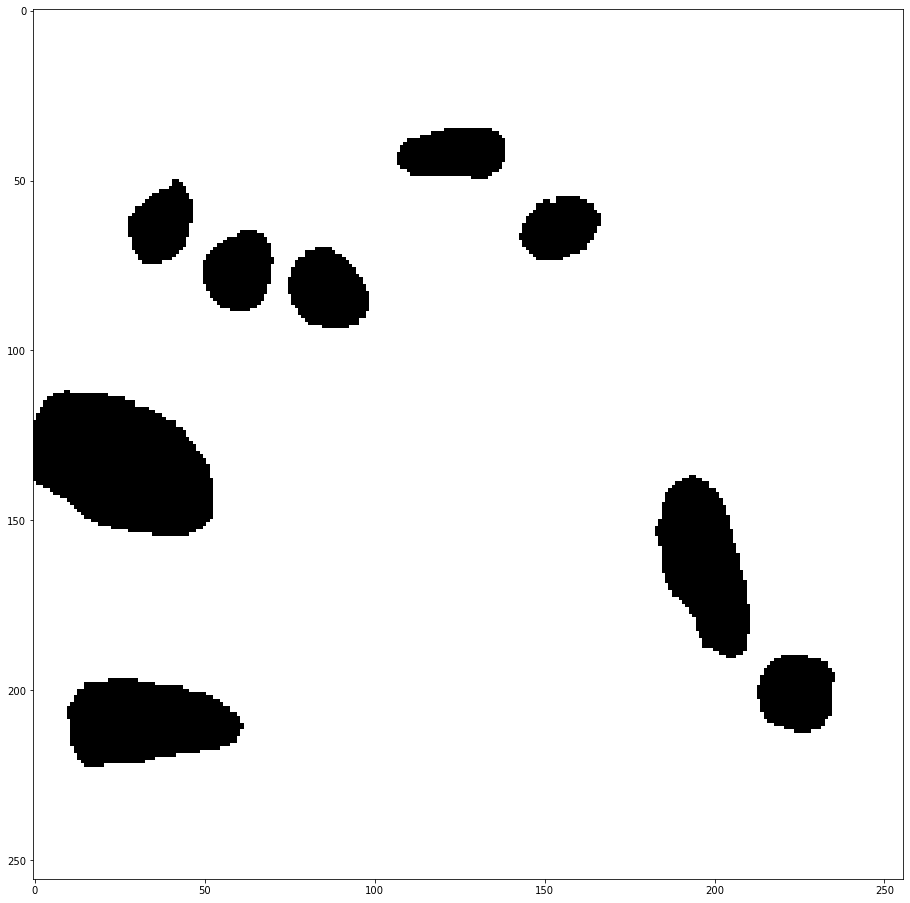}
            \caption{Masks with detection errors}
            \label{fig:detection_error}
            \end{subfigure}
            \begin{subfigure}[t]{0.3\textwidth}
            \includegraphics[width=\linewidth]{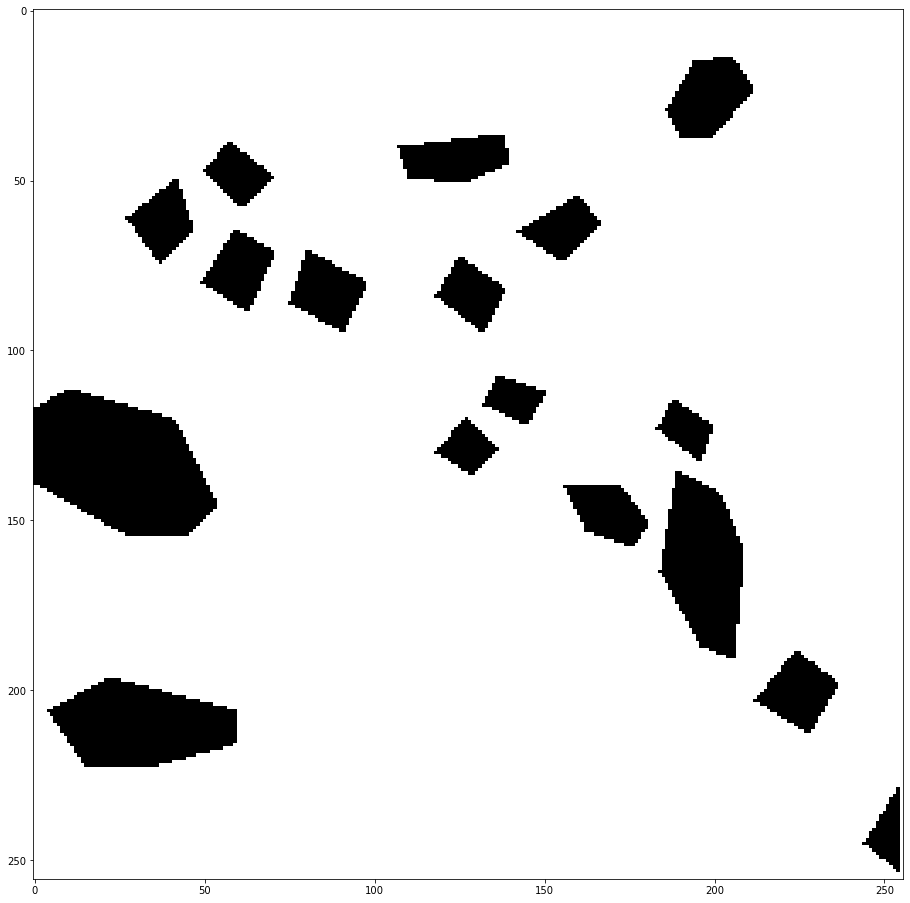}
            \caption{Masks with segmentation errors}
            \label{fig:segmentation_error}
            \end{subfigure}
            \begin{subfigure}[t]{0.3\textwidth}
            \includegraphics[width=\linewidth]{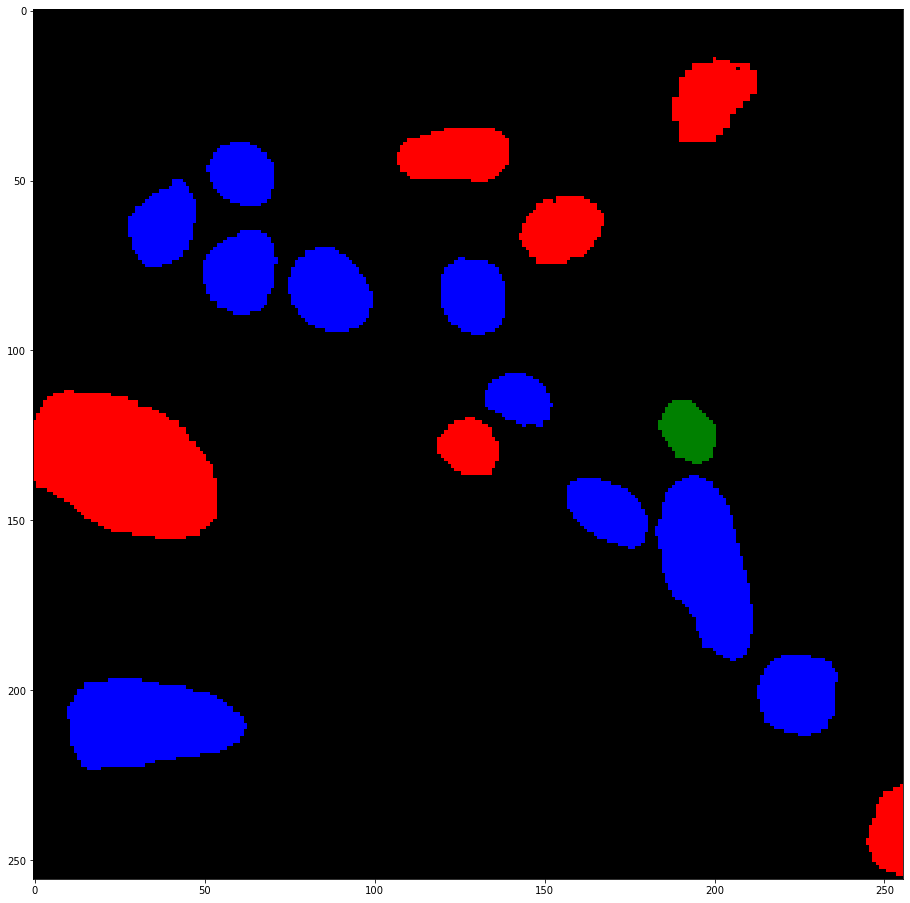}
            \caption{Masks with classification errors}
            \label{fig:classification_error}
            \end{subfigure}
            \caption{(a-c) Examples of annotations and (d-f) their corruption introduced in the clean MoNuSAC training subset. The classification masks show epithelial cells in red, lymphocytes in blue and neutrophils in green.}
            \label{fig:errors_example}
            \end{figure*}

Annotating large datasets is a time-consuming and complex task that leads to human errors. For large numbers of small objects, such as cell nuclei, the most common errors include missing objects, inaccurate delineations, and/or confusion in assigning class labels in multi-class datasets.
To mimic these errors, we introduce three types of realistic annotation corruptions in clean training sets impacting detection, segmentation, and classification, respectively, as illustrated in Figure~\ref{fig:errors_example}.

To simulate detection and classification errors we adapt the Noisy at Random Model described by Frénay and Verleysen \cite{frenay2013classification}, where the occurrence of a label error depends on the true class but is independent of the features of the observed annotation. This model is characterised by the following noise transition matrix:
\begin{equation}
    Q_{i,j}=P(\widetilde{Y_l}=i|Y_l=j) 
    \label{eq:Q_general}
\end{equation}
where, in our case, $\widetilde{Y_l}$ and $Y_l$ are respectively the corrupted and the original label of any pixel region $R_l$ (possibly an object of interest); $i$ and $j$ can take values from 0 to $K$, 0 corresponding to the background, and $K$ being the number of classes of interest. As detailed below, the background is only involved in detection errors.

\begin{description}

    \item[Detection errors.]
    When numerous objects of interest are present in an image, they are often not all detected and thus annotated, resulting in the absence of segmentation and classification masks for these objects.
    While it is easy to mimic this common error by removing annotations, adding new annotations in the background in a realistic way, which would require the identification of objects that resemble cell nuclei but are not, is challenging.
    We thus consider only asymmetrical noise: 
    $P(\widetilde{Y_l}=i|Y_l=0;i>0)=0$. 
    We randomly remove a certain percentage $\rho$ of the nucleus annotations for each class $i$ (i.e., without modifying the class priors), so that $P(\widetilde{Y_l}=0|Y_l=i;i>0)=\rho$. 
    These corruptions can be summarised by the following 2x2 noise transition matrix independently applied to each class:
 \begin{equation}
        Q^D_\rho = 
        \begin{pmatrix}
            1 & \rho\\
            0 & 1-\rho\\
        \end{pmatrix},
    \end{equation}
   where the first row and column concern the background and the second row and column a given class.

    The removed nuclei are relabelled as background by setting their pixel values to $0$ in both classification and segmentation masks. This process is illustrated in Figure \ref{fig:detection_error}, where 20\% of the nucleus annotations are randomly removed.
            
    \item[Segmentation errors.]
    Although the nuclei are accurately detected, their contours may not be precisely delineated.
    This discrepancy could result from the simplification of the annotation task through the use of bounding boxes or polygons as a convenient approximation of the actual nuclei contours.
    To introduce imprecise contours, we simplify the original annotations into polygons with a reduced number of sides, as depicted in Figure \ref{fig:segmentation_error}.
    For each annotated nucleus $l$, a two-step process is employed.
    
    First, we fit an ellipse $E_l(\Vec{a})$, parameterised by vector $\Vec{a}$, to the set of pixels defining the nucleus contour, $C_l$, using the algebraic distance algorithm \cite{fitzgibbon1996buyer}.
    Second, a simplification factor is applied to obtain an ellipse $E_{l'}(\Vec{a})$, from which we extract an approximate polygon $l'$ using the Douglas-Peucker algorithm \cite{douglas1973algorithms}, with a tolerance $\epsilon$ chosen to replicate realistic segmentation errors. Bounding boxes are not employed in this step due to the highly variable orientations of the nuclei and their density.
    
    Moreover, segmenting adjacent and overlapping nuclei poses a challenge, potentially leading to over-segmentation or, more commonly, under-segmentation, particularly in human annotations of small objects like cell nuclei. To simulate this prevalent issue of under-segmentation, we merge the masks of pairs of adjacent nuclei of the same class. This allows us to distinguish the effects of under-segmentation from those of classification errors. The process involves assigning the same instance ID to merged nuclei and subsequently smoothing the contours for a more realistic appearance. In cases where more than two adjacent nuclei are present, we merge the pair with the largest common border area.

    \item[Classification errors.]
    While all nuclei are successfully detected and segmented, not all of them are accurately assigned to the correct class. This discrepancy can arise from the apparent similarity between certain cell types or variations in the annotator's domain knowledge and expertise.
    Unlike detection noise, we introduce classification errors as symmetric noise using the noise transition matrix $Q_{i,j}$ in equation (\ref{eq:Q_general}) but with $i,j>0$ (since the background is not considered), resulting in the following ($K \times K$) matrix:
    \begin{equation}
        Q^C_\rho = 
        \begin{pmatrix}
            1-\rho & \dfrac{\rho}{K-1} & \cdots & \dfrac{\rho}{K-1}\\
            \dfrac{\rho}{K-1} & 1-\rho &  \cdots & \dfrac{\rho}{K-1}\\
            \vdots & \vdots & \ddots & \vdots \\
            \dfrac{\rho}{K-1}  & \dfrac{\rho}{K-1} & \cdots & 1-\rho\\
        \end{pmatrix}
    \end{equation}
     This leads to a random exchange of labels for the same percentage ($\rho$) of nuclei in each class, as depicted in Figure \ref{fig:classification_error}, where the labels for 20\% of the nuclei in each class are altered.
\end{description}   

These error types are incrementally introduced into the training set in increasing proportions and can also be combined, as detailed in Section \ref{sec:results}.

\subsection{Model and training process} \label{section:Model}
To analyse the impact of imperfect annotations on cell nuclei, we employ the Hover-Net model \cite{graham2019hover}. We specifically selected this architecture for the following reasons: (i) it secured the top position in the MoNuSAC challenge ranking \cite{verma2021monusac2020}, demonstrating, in particular, its robustness to imperfect training set annotations, as highlighted in \cite{galvez2023cleaning}, and (ii) it has been recognised as one of the best models for cell nuclei segmentation and classification in recent literature \cite{doan2022sonnet,gudhe2023nuclei,wang2023improved}. Another advantageous feature of Hover-Net in the context of this study is its seamless integration of a pre-training step into the training algorithm, as detailed below.

The network comprises two parts: the encoder for feature extraction and the decoder for instance segmentation and classification \cite{graham2019hover}. Briefly, the first part is based on a pre-activated residual network with 50 layers (Preact-ResNet50), while the second part uses three specialised branches to segment and classify simultaneously, as follows. The Nuclear Pixel (NP) branch predicts whether a pixel belongs to the background or a nucleus; the HoVer (HV) branch predicts the horizontal and vertical distances of nuclear pixels from their centres of mass; the Nuclear Classification (NC) branch predicts the type of nucleus for each pixel. The final output is an instance segmentation and classification mask which combines the resulting instance segmentation map, obtained from the NP and HV branches, and the classification map obtained from the NC branch. In the results, we identify the objects that the network detects as nuclei without being able to classify them into one of the classes of interest by assigning them to an "other" class, to distinguish between classification errors and detection errors. 

We use the HoVer-Net loss proposed in the original paper \cite{graham2019hover} that is defined as the combination of the losses of the three branches $L = L_{HV} + L_{NP} + L_{NC}$, where $L_{HV}$ is the HoVer branch loss calculated as a regression loss and $L_{NP}$ and $L_{NC}$ are the NP and NC branch losses, both calculated as a combination of the cross-entropy loss and the dice loss. Additional details can be found in the original paper.

For all the experiments, we train the network following the methodology employed by the TIA-lab team, which led to their success in winning the MoNuSAC challenge \cite{verma2021monusac2020}. As illustrated at the top of the flowchart in Figure \ref{flowchart} in the Appendix, the complete network is initialised with pre-trained weights on Imagenet and then trained on a task similar to the target task, utilizing a first clean training set to refine the results. The model is subsequently trained on the target task using a second training set. In most experiments, data from PanNuke serves as training set 1, and data from MoNuSAC (V1.2) serves as training set 2. Their roles are then reversed in the final experiments for validation purposes. Pre-trained weights with Imagenet and PanNuke are publicly available\footnote{The original implementation is available from the authors at \url{https://github.com/vqdang/hover_net}}.

For each training set (1 or 2) mentioned above, the training process is divided into two stages, as described in Algorithm \ref{hover_training}: (i) the first stage freezes the (pre-trained) encoder part and trains only the convolutional layers of the three specialised branches (NP, HV and NC), and (ii) the second stage fine-tunes the entire network, taking as initial weights those from the first stage. Following the study by Verma et al. \cite{verma2021monusac2020}, the maximum number of epochs is set to 50 for the first and second stages, both implemented with Adam optimisation and an initial learning rate of $10^{-4}$, which is reduced to $10^{-5}$ in the second stage. 

In addition, we introduce training-stopping conditions, as explained in the next section (see also Figure \ref{flowchart}). 

\begin{algorithm}[!ht]
    \caption{HoVer-Net training adapted from \cite{graham2019hover, verma2021monusac2020}}\label{hover_training}
    \begin{algorithmic}
        \State \textbf{First stage}
        \Require Pre-trained weights $M_0$, maximum number of epochs $N$and freeze encoder. 
        \State $M_e \gets M_0$
        \State $e \gets 0$
        \While{$!EarlyStoppingCondition$ \& $e < N$}
            \State Train model in epoch $e$ to get new $M_e$
            \State Evaluate model $M_e$ on a validation set
            \State Apply early stopping procedure to get $M_{best}$
        \EndWhile
        
        \State \textbf{return} $M_{best}$
        \\
        \State \textbf{Second stage}
        \Require Pre-trained weights $M_{best}$, maximum number of epochs $N{bis}$ and unfreeze encoder.
        \State $M_{bis_e} \gets M_{best}$
        \State $e \gets 0$
        \State $M_{bis_{best}} \gets M_{best}$ 
        \While{$!EarlyStoppingCondition$ \& $e < N{bis}$}
            \State Train model in epoch $e$ to get new $M_{bis_e}$
            \State Evaluate model $M_{bis_e}$ on a validation set
            \State Apply early stopping procedure to get $M_{bis_{best}}$
        \EndWhile
        
        \State \textbf{return} $M_{bis_{best}}$
    \end{algorithmic}
\end{algorithm}

\subsection{Training stopping in the presence of noisy datasets}
\label{sec:TrainingStopping}

To determine the number of training epochs and avoid overfitting on noisy annotations, we analyse the HoVer-Net loss computed on a validation set.

The same stop-training method is applied in both stages of model formation. The goal is to identify the epoch at which the model starts exhibiting training faults indicated by a tendency for the loss to increase (see Algorithm \ref{early_stopping_procedure} and Figure \ref{flowchart}). 
This procedure involves selecting suitable parameters: a patience value, $p$ (the number of epochs without improvement before stopping training), and a minimum loss variation, $\delta$, to consider improvements.
The values for these parameters are determined experimentally based on training with a high number of epochs, allowing for the observation of the fluctuation range of the loss. Examples are provided in Section \ref{sec:experiments_stopping}.

\begin{algorithm}
    \caption{Model training stopping}\label{early_stopping_procedure}
    \begin{algorithmic}
        \Require Set patience $p$, minimum value variation $\delta$ and maximum number of epochs $N$.
        \State $c \gets 0$
        \State $e \gets 0$
        \State $M_e \gets M_0$
        \State $M_{best} \gets M_0$
        \State $l_{min} \gets \infty$
        \While{$c < p$ \& $e < N$}
            \State $l_e \gets loss(M_e)$
            \If{$l_e < (l_{min}+\delta)$} 
                \State $l_{min} \gets l_e$
                \State $c \gets 0$
                \State $M_{best} \gets M_e$
            \Else
                \State $c \gets c+1$
            \State $e \gets e+1$
            \EndIf
        \EndWhile
        
        \textbf{return} $M_{best}$
    \end{algorithmic}
\end{algorithm}

Regarding the MoNuSAC dataset, to ensure a validation set with as accurate annotations as possible, we extract a small but representative subset from the test set (which is well-annotated, see Section \ref{section:datasets}). Specifically, we select one patient per tissue type, resulting in a subset comprising 4 of the 25 patients constituting the test set. To prevent biases linked to this selection, we repeat this process 5 times, yielding 5 different pairs of complementary validation and test subsets.
Concerning the PanNuke dataset, from the fold designated for test and validation, we generate 5 different pairs of subsets with 80\% of the patches from each tissue type allocated to the test set and 20\% for the validation {set}.

To assess the impact of the annotation quality of the validation set on determining an appropriate number of training epochs, we conduct another experiment in which we extract a small portion of the training set with corrupted annotations to serve as the validation set. Consequently, the annotation noise present in the training set is transmitted to the validation set.

\subsection{Performance metrics} \label{sec:metrics}
We use the performance metrics described in previous works \cite{galvez2023cleaning,foucart2022evaluating} since the Panoptic Quality used in the MoNuSAC challenge is considered flawed, making evaluation challenging to interpret \cite{foucart2022evaluating,foucart2022comments}. The metrics assess each of the detection, segmentation, and classification tasks separately. This implies that segmentation and classification performance are evaluated only on correctly detected nuclei, as follows:

\begin{description}
\item[Detection:] To identify the ``well-detected'' nuclei, we utilise the matching rule proposed by Foucart et al. \cite{foucart2022evaluating}. Among all possible pairs of annotated and predicted objects, the rule selects the pair with the highest intersection over union (IoU) and considers it an effective match if the predicted instance centroid belongs to the annotation. Based on this, we determine matching pairs as well as unmatched predictions (i.e., false positives) and unmatched annotated instances (i.e., false negatives) to quantify precision, recall, and F1-score.

\item[Segmentation:] Segmentation performance is evaluated by calculating the IoU and the Hausdorff Distance (HD) between each well-detected nucleus and its matched annotated nucleus.
In addition, we calculate the number of over- and under-segmentation cases. 
Cases of over-segmentation are identified as detected objects whose surface predominantly belongs to a ground truth instance (with an IoU of at least 0.5) that is already matched to another detected nucleus. These objects contribute to false-positive detections. 
Conversely, cases of under-segmentation are identified as ground truth instances whose surface predominantly belongs to a detected nucleus (with an IoU of at least 0.5) that is already matched to another ground truth instance. These missing nuclei contribute to false-negative detections. 

\item[Classification:] We assess the classification performance on well-detected nuclei by computing different balanced metrics from the Normalised Confusion Matrix (NCM), for which the sum on each row, relative to a true class, is set to 1 or 100\%. These metrics include accuracy, precision (or specificity), recall (or sensitivity), and F1-score per class (see details in Section \ref{detection_noise_results}).
\end{description}

Detection and segmentation performance are class-independent. However, it is interesting to detail the contribution of each class to the overall results, given the strong imbalance of classes in the datasets used.

%% file: results.tex
\section{Experiments and results}
\label{sec:results}

In the following experiments, we introduce the different types of annotation errors described in Section \ref{section:imperfections}, which we refer to as "noise", to obtain corrupted versions of the clean V1.2 MoNuSAC training set. We first analyse the impact of annotation quality in the validation set on the training stopping strategy described in Section \ref{sec:TrainingStopping}. Then, we assess the impact of annotation noise in the training set on the model performance measured on the original test set, with or without training stopping. In the last experiments, we reverse the roles played by the MoNuSAC and PanNuke datasets in the training process (i.e., pre-training with V1.2 MoNuSAC and then training with PanNuke with or without noisy annotations), as described in Section \ref{section:Model}.

All experiments are conducted according to the flowchart in Figure \ref{flowchart}.

\subsection{Impact of annotation noise on the ability to stop training appropriately}
\label{sec:experiments_stopping}

\begin{figure*}[!ht]
            \centering
            \begin{subfigure}[t]{0.48\textwidth}
            \includegraphics[width=\linewidth]{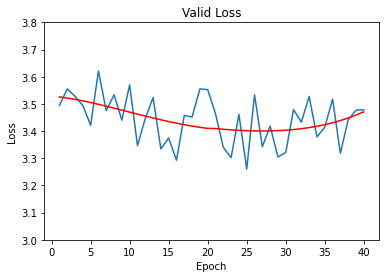}
            \caption{1st training stage with clean training and validation sets}
            \label{fig:loss_clean}
            \end{subfigure}
            \begin{subfigure}[t]{0.48\textwidth}
            \includegraphics[width=\linewidth]{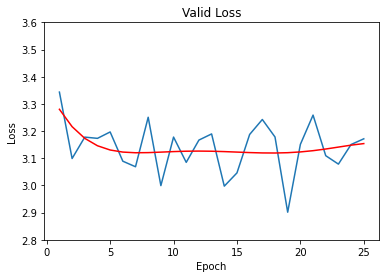}
            \caption{1st training stage with segmentation noise, clean validation}
            \label{fig:loss_seg}
            \end{subfigure}
            \begin{subfigure}[t]{0.48\textwidth}
            \includegraphics[width=\linewidth]{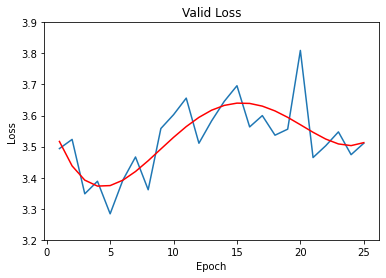}
            \caption{1st training stage with 40\% detection noise, clean validation}
            \label{fig:loss_det}
            \end{subfigure}
            \begin{subfigure}[t]{0.48\textwidth}
            \includegraphics[width=\linewidth]{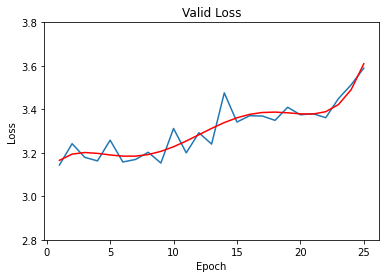}
            \caption{1st training stage with 40\% detection noise, noisy validation}
            \label{fig:loss_det_noisy1}
            \end{subfigure}
            \begin{subfigure}[t]{0.48\textwidth}
            \includegraphics[width=\linewidth]{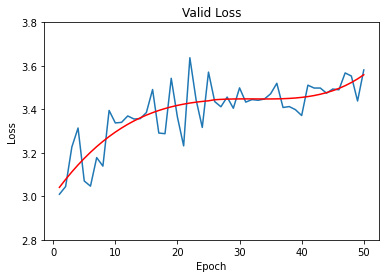}
            \caption{2nd training stage with 40\% detection noise, noisy validation}
            \label{fig:loss_det_noisy2}
            \end{subfigure}
            \caption{Loss obtained when evaluating Algorithm \ref{hover_training} training stages} using clean or noisy validation sets. The raw curves are in blue and their smooth versions in red. (a) First stage of training with the clean V1.2. training set, evaluated on a clean validation set. (b) First stage of training with segmentation noise in the training set, evaluated on a clean validation set. (c) First stage of training with 40\% detection noise, evaluated on a clean validation set. (d) First and (e) second stage of training with 40\% detection noise, both evaluated on a noisy validation set (40\% detection noise).
            \label{fig:early_stopping}
            \end{figure*}
            
Using the MoNuSAC dataset, we investigate the training stopping strategy described in Section \ref{sec:TrainingStopping} in the presence of various types and levels of annotation noise introduced in the training set, as defined in Section \ref{section:imperfections}. Figure \ref{fig:early_stopping} illustrates the loss curves computed on a validation set during HoVer-Net's training (via Algorithm \ref{hover_training}) with MoNuSAC.
To better visualise the trends, we add smooth curves obtained using the Savitzky-Golay filter with a fourth-order polynomial to fit the loss values. Examining the curves, their trends and fluctuations, in Algorithm \ref{early_stopping_procedure} we set patience $p$ to 10 and $\delta$ to 0.001 for the first experiment (Figure \ref{fig:loss_clean}), changing $\delta$ to 0.01 for the second and third experiments (Figures \ref{fig:loss_seg}-\ref{fig:loss_det}) to determine the optimal stopping point (see Section \ref{sec:TrainingStopping}). 

In the case of the clean V1.2 training set, the best epochs determined by Algorithm \ref{early_stopping_procedure} correspond to 25 for the first training stage of Algorithm \ref{hover_training} (Figure \ref{fig:loss_clean}) and 50 for the second. After introducing segmentation noise into this training set, the loss curve computed during the first training stage of Algorithm \ref{hover_training} on a clean validation set suggests stopping training earlier, identified by Algorithm \ref{early_stopping_procedure} as epoch 19 (see Figure \ref{fig:loss_seg}). Even more obviously, with the introduction of 40\% detection noise in the training set, the loss curve computed on a clean validation set during the first training stage also indicates the need to stop training prematurely, with epoch 5 identified by Algorithm \ref{early_stopping_procedure} as optimal before the loss increases (see Figure \ref{fig:loss_det}). 
Similar behaviours are observed during the second training stage of Algorithm \ref{hover_training}. The number of epochs output by Algorithm \ref{early_stopping_procedure} under different training conditions and for both training stages are detailed in the next section.

It is customary to extract a small part of the training set as a validation set to determine the number of epochs. The annotation noise present in the training set is therefore transmitted to the validation set. Figures \ref{fig:loss_det_noisy1} and \ref{fig:loss_det_noisy2} depict the loss curves obtained for the two training stages of Algorithm \ref{hover_training} when we extract a small validation set from a training set with 40\% detection noise.

During the first training stage, the loss curve shows that it is also necessary to stop the training earlier than in the case of a clean training set (see Figure \ref{fig:loss_det_noisy1}). Indeed, Algorithm \ref{early_stopping_procedure} (setting patience $p$ to 10 and $\delta$ to 0.01) identifies epoch 9 as the optimal point to stop training, which is later than the epoch obtained using a clean validation set. 
Furthermore, during the second training stage, the loss curve increases from the first epoch, which would be the stopping point if we apply Algorithm \ref{early_stopping_procedure} (see Figure \ref{fig:loss_det_noisy2}). However, the model obtained with these numbers of training epochs (9 and 1 for the first and second stage, respectively) provides poorer performance on the test set than that obtained 
when training stopping is based on a clean validation set (described in the next section). 
In particular, the neutrophil class experiences a decrease of at least 5\% in both detection and classification F1-scores. Moreover, the balanced accuracy is even lower than that obtained without applying early training stopping.  
Given the above results, the experiments below focus on analysing the impact of annotation noise on Hover-Net's performance, with and without the use of our early stopping strategy based on a clean validation set.

\subsection{Impact of detection noise on Hover-Net performance}
\label{detection_noise_results}

We introduce varying levels of detection noise, low (20\%), medium (40\%) and high (60\%), into the annotations of the clean V1.2 MoNuSAC training set, to observe the model's robustness to these imperfections for each task (detection, segmentation, and classification). 
For each level of noise, we compare the performance of models trained using the number of epochs determined from a clean validation against those trained with 25 epochs for the first training stage and 50 epochs for the second, as employed by the reference model trained with clean training and validation sets. This comparison aims to show the impact that overfitting to annotation noise would have on the model's performance.

For a better explanation of the results obtained for the detection and classification metrics detailed below, we present in Table \ref{Tab:cmDatasetVersions} the confusion matrices (CMs) obtained on one test set with different models and training conditions. Matrices (a)-(c) detail the raw results, and matrices (d)-(f), labelled NCMs, correspond to their normalised values used to extract the balanced classification metrics. 

To assess the detection task, we have to consider the "U" column and the "$\emptyset$" row of the raw CMs. While the "U" column refers to undetected nuclei by the network (i.e., false negatives), the "$\emptyset$" row corresponds to the actual background as well as unknown/unannotated nuclei, which are nevertheless subject to detection and classification by the network. The last three cells of this row, therefore, provide what we consider as false detections (i.e., false positives), detailed for classes E, L, and N.

The introduction of detection noise in the training set means that the model encounters larger background regions without annotated nuclei (compared to the clean training set). The raw CMs show that this results in an increase in the number of undetected nuclei. However, the use of early training stopping keeps this increase at a much lower level (overall multiplicative factor of 1.58, instead of 3.46 observed without training stop), while increasing the number of false detections but still having a positive impact on the F1 score (see below).

\begin{table*}[!ht]
    \centering
    \caption{Predictions on a test set that are provided by three Hover-Net models trained with the clean V1.2 training set (a and d) or its corrupted version with 40\% detection noise, with (b and e) or without (c and f) early training stopping. (a-c) The raw CMs include an $\emptyset$ line (background or unknown/unannotated nuclei), an ``U'' column (undetected nuclei) and an "other" column (detected but not classified nuclei). (d-f) The normalised CMs (expressed as percentages) are computed only on the well-detected nuclei for extracting balanced classification metrics. E, L, N designate epithelial, lymphocyte and neutrophil classes respectively.}

\makebox[\textwidth]{
\setcounter{subtable}{0}
\begin{subtable}{0.49\textwidth}
\centering
    \begin{tabular}{cc|ccccc}
     & & \multicolumn{5}{c}{Predicted} \\ 
     & & U & Other & E & L & N \\
    \cline{2-7}
    \multirow[c]{4}{*}{\rotatebox[origin=tr]{90}{Actual}}
    & $\emptyset$ & ? & 124 & 757 & 569 & 16 \\
    & E & 874 & 22 & 4973 & 57 & 2 \\
    & L & 440 & 18  & 157 & 5311 & 14 \\
    & N & 21 & 0 & 2 & 3 & 105 \\ 
    \cline{2-7}
    \end{tabular}
    \caption{Raw CM - Clean training set}
\end{subtable}

\hfill

\setcounter{subtable}{3}
\begin{subtable}{0.49\textwidth}
\centering
\begin{tabular}{cc|cccc}
     & & \multicolumn{4}{c}{Predicted} \\ 
     & & Other & E & L & N \\
    \cline{2-6}
    & & & & & \\
    & E & 0.4 & 98.4 & 1.1 & 0.0 \\
    & L & 0.3 & 2.9 & 96.6 & 0.2 \\
    & N & 0.0 & 1.8 & 2.7 & 95.5 \\
    \cline{2-6}
    \end{tabular}
    \caption{NCM - Clean training set}
\end{subtable}
}
\makebox[\linewidth]{
\setcounter{subtable}{1}
\begin{subtable}{0.49\textwidth}
\centering
    \begin{tabular}{cc|ccccc}
     & & \multicolumn{5}{c}{Predicted} \\ 
     & & U & Other & E & L & N \\
    \cline{2-7}
    \multirow[c]{4}{*}{\rotatebox[origin=tr]{90}{Actual}}
    & $\emptyset$ & ? & 237 & 925 & 655 & 41 \\
    & E & 1347 & 55 & 4454 & 70 & 2 \\
    & L & 733 & 21 & 57 & 5114 & 15 \\
    & N & 34 & 0 & 0 & 8 & 89 \\ 
    \cline{2-7}
    \end{tabular}
    \caption{Raw CM - Noise, early-stopping}
\end{subtable}

\hfill
\setcounter{subtable}{4}
\begin{subtable}{0.49\textwidth}
\centering
\begin{tabular}{cc|cccc}
     & & \multicolumn{4}{c}{Predicted} \\ 
     & & Other & E & L & N \\
    \cline{2-6}
    & & & & & \\
    & E & 1.2 & 97.2 & 1.5 & 0.0 \\
    & L & 0.4 & 1.1 & 98.2 & 0.2 \\
    & N & 0.0 & 0.0 & 8.2 & 91.8 \\
    \cline{2-6}
    \end{tabular}
    \caption{NCM - Noise, early-stopping}
\end{subtable}
}

\makebox[\linewidth]{
\setcounter{subtable}{2}
\begin{subtable}{0.49\textwidth}
\centering
    \begin{tabular}{cc|ccccc}
     & & \multicolumn{5}{c}{Predicted} \\ 
     & & U & Other & E & L & N \\
    \cline{2-7}
    \multirow[c]{4}{*}{\rotatebox[origin=tr]{90}{Actual}}
    & $\emptyset$ & ? & 321 & 607 & 434 & 15 \\
    & E & 3065 & 81 & 2742 & 38 & 2 \\
    & L & 1498 & 163 & 130 & 4144 & 5 \\
    & N & 51 & 5 & 1 & 4 & 70 \\ 
    \cline{2-7}
    \end{tabular}
    \caption{Raw CM - Noise, no early-stopping}
\end{subtable}

\hfill
\setcounter{subtable}{5}
\begin{subtable}{0.49\textwidth}
\centering
\begin{tabular}{cc|cccc}
     & & \multicolumn{4}{c}{Predicted} \\ 
     & & Other & E & L & N \\
    \cline{2-6}
    & & & & & \\
    & E & 2.8 & 95.8 & 1.3 & 0.1 \\
    & L & 3.7 & 2.9 & 93.3 & 0.1 \\
    & N & 6.3 & 1.3 & 5.0 & 87.5 \\
    \cline{2-6}
    \end{tabular}
    \caption{NCM - Noise, no early-stopping}
\end{subtable}
}
\label{Tab:cmDatasetVersions}
\end{table*}  

Table \ref{Tab:DetectMetricsDetectionNoise} provides a detailed breakdown of the detection metrics computed on the five test sets, along with the determined numbers of training epochs based on the corresponding clean validation sets.
The results underscore a gradual decline in the model's performance as the level of detection noise increases. The introduction of a medium level of detection noise into the training set markedly reduces the overall F1-score. However, the application of our early training stopping strategy results in improvement.
Consistent with our observations in the CMs, precision diminishes when employing this strategy due to an increase in false detections. In contrast, recall improves substantially as the number of detected nuclei increases, leading to substantially improved F1-scores.
The diverse behaviour among classes may stem from their distinct features. Epithelial cells, for instance, may exhibit varying appearances across different organs. Neutrophils, being rarer, can change shape based on their angle of rotation, while lymphocytes, generally rounded cells, are likely easier to detect.
 
\begin{table*}[!ht]
    \centering
    \caption{Impact of different levels of detection noise introduced in the training set on the detection metrics computed on the test set, for models obtained with or without early training stopping. In the case of early training stopping, the number of epochs is indicated by the range of values obtained on 5 validation sets. Results on 5 complementary test sets are expressed in percentages as mean(std).}
    
    \begin{tabular}{@{}cc|c|cc|cc|cc@{}}
    \multicolumn{1}{c}{} &\multicolumn{1}{c}{} &\multicolumn{4}{c}{} \\ 
    \multicolumn{2}{c|}{\textbf{Detection}} & 
    \multicolumn{1}{c|}{V1.2 (clean)} &
    \multicolumn{2}{c|}{20\% detection noise} &
    \multicolumn{2}{c|}{40\% detection noise} &
    \multicolumn{2}{c}{60\% detection noise}\\

    \hline
    \multirow[c]{2}{*}{{Epochs}}
    & 1st stage &  25 & 14-17 & 25 & 5 & 25 & 1 & 25\\
    & 2nd stage &  50 & 4-6 & 50 & 3-4 & 50 & 1 & 50\\
    \hline
    \multirow[c]{4}{*}{{Precision}}
    & E & 84.3(0.8) & 83.7(0.4) & 83.8(0.7) & 82.0(0.5) & 79.6(0.5) & 83.3(0.7) & 79.9(0.5)\\
    & L & 88.3(0.8) & 84.0(1.1) & 89.7(0.7) & 86.5(0.6) & 89.4(0.5) & 83.4(0.7) & 82.9(0.6)\\
    & N & 80.0(2.2) & 71.0(2.9) & 83.6(3.1) & 66.3(4.2) & 79.6(2.0) & 57.6(2.6) & 75.7(6.4)\\
    & \textbf{Overall} & 87.3(0.5) & 85.1(0.4) & 87.2(0.4) & 83.7(0.3) & 84.2(0.4) & 83.4(0.4) & 79.2(0.3)\\ 

    \hline
    \multirow[c]{4}{*}{{Recall}}
    & E & 85.9(0.7) & 86.4(0.5) & 76.7(1.2) & 78.3(0.9) & 50.6(1.6) & 73.0(1.1) & 28.0(0.7) \\
    & L & 92.9(0.3) & 92.9(0.3) & 89.3(0.6) & 88.0(0.5) & 76.4(0.9) & 82.5(0.5) & 41.3(1.1) \\
    & N & 85.9(0.4) & 91.3(1.1) & 79.1(2.8) & 73.3(1.8) & 57.6(2.3) & 63.1(2.5) & 28.3(3.1) \\ 
    & \textbf{Overall} & 89.5(2.3) & 89.8(0.3) & 83.2(0.9) & 83.2(0.6) & 63.9(1.4) & 77.8(0.6) & 34.8(0.7) \\ 

    \hline
    \multirow[c]{4}{*}{{F1-Score}}
    & E & 85.1(0.5) & 85.1(0.3) & 80.1(0.7) & 80.1(0.4) & 61.8(1.2) & 77.8(0.8) & 41.4(0.9) \\
    & L & 90.6(0.3) & 88.2(0.6) & 89.5(0.2) & 87.2(0.3) & 82.4(0.4) & 82.9(0.3) & 55.2(1.0) \\
    & N & 82.8(1.8) & 79.9(1.9) & 81.3(2.7) & 70.0(2.5) & 66.8(0.8) & 60.2(1.8) & 41.2(4.1) \\ 
    & \textbf{Overall} & 88.4(0.3) & 87.4(0.2) & 85.1(0.5) & 83.7(0.2) & 72.7(1.0) & 80.5(0.5) & 48.4(0.7) \\ 
    
    \end{tabular}
    
    \label{Tab:DetectMetricsDetectionNoise}
\end{table*}

Table \ref{Tab:DetectMetricsDetectionNoise} also highlights that the number of training epochs determined with a clean validation set decreases as the level of annotation noise increases in the training set. Consequently, a higher level of annotation noise necessitates earlier training interruption to prevent noise overfitting. 

In addition, Table \ref{Tab:SegMetricsDetectionNoise} illustrates that segmentation performance is impacted by the introduction of detection noise, with a more noticeable effect starting from 40\% noise. Epithelial cells, in particular, are the most noticeably impacted in the absence of early training stopping, potentially due to the variability in their appearance, as mentioned earlier. These cells benefit the most from early training stopping.

Moreover, Table \ref{Tab:over_under_seg} in the Appendix confirms that with early training stopping, the introduction of 40\% detection noise in the training set increases false-positive detections and, notably, false-negative detections, with only a minor increase in under-segmentation problems. Without early stopping, the number of under-segmentation instances is reduced by around 50\%, but at the cost of a significant increase in the number of false negatives.

\begin{table*}[!ht]
    \centering
    \caption{Impact of different levels of detection noise introduced in the training set on the segmentation metrics computed on the test set, for models with or without early training stop. The results are expressed as in Table \ref{Tab:DetectMetricsDetectionNoise}, except for HD, which is expressed in pixels.} 
    \begin{tabular}{@{}cc|c|cc|cc|cc@{}}
    \multicolumn{1}{c}{} &\multicolumn{1}{c}{} &\multicolumn{4}{c}{} \\ 
    \multicolumn{2}{c|}{\textbf{Segmentation}} & 
    \multicolumn{1}{c|}{V1.2 (clean)} &
    \multicolumn{2}{c|}{20\% detection noise} &
    \multicolumn{2}{c|}{40\% detection noise} &
    \multicolumn{2}{c}{60\% detection noise}\\ 
    \hline
    \multirow[c]{2}{*}{{Epochs}}
    & 1st stage &  25 & 14-17 & 25 & 5 & 25 & 1 & 25\\
    & 2nd stage &  50 & 4-6 & 50 & 3-4 & 50 & 1 & 50\\
    \hline
    \multirow[c]{4}{*}{{IoU}}
    & E & 80.8(0.3) & 80.2(0.3) & 78.5(0.2) & 78.0(0.2) & 73.1(0.2) & 73.1(0.5) & 70.1(0.6) \\
    & L & 76.6(0.2)& 74.8(0.2) & 75.7(0.3) & 71.8(0.2) & 72.3(0.3) & 68.8(0.3) & 66.9(0.3) \\
    & N & 76.5(0.6) & 77.5(0.8) & 76.0(1.0) & 76.5(0.7) & 74.2(1.2) & 65.6(1.4) & 72.0(1.6) \\ 
    & \textbf{Overall} & 78.0(0.2) & 77.5(0.2) & 76.7(0.3) & 75.5(0.1) & 73.2(0.3) & 69.2(0.3) & 69.7(0.4) \\
    \hline
    \multirow[c]{4}{*}{{HD}}
    & E & 4.7(0.1) & 5.0(0.1) & 5.2(0.1) & 5.5(0.1) & 7.0(0.2) & 6.3(0.2) & 8.4(0.1) \\
    & L & 3.0(0.0) & 3.3(0.0) & 3.1(0.0) & 3.5(0.1) & 3.5(0.0) & 3.8(0.1) & 4.2(0.1) \\
    & N & 4.3(0.0) & 4.2(0.2) & 3.9(0.2) & 4.3(0.3) & 5.2(0.3) & 7.2(0.2) & 5.7(0.4) \\ 
    & \textbf{Overall} & 3.7(0.1) & 3.9(0.1) & 4.0(0.1) & 4.3(0.1) & 4.6(0.1) & 4.8(0.1) & 5.5(0.1) \\

    \end{tabular}
    
    \label{Tab:SegMetricsDetectionNoise}

\end{table*}

Classification performance is assessed based on the raw CM elements that are not considered for detection (i.e. excluding the U column and the $\emptyset$ line), from which NCM is calculated (see Table \ref{Tab:cmDatasetVersions}).
It is important to note that the percentages in NCM refer to the numbers of well-detected nuclei, which strongly vary across different training conditions (as detailed in  Table \ref{Tab:ClassMetricsDetectionNoise}).
In contrast to detection performance, the computed classification metrics remain relatively stable regardless of the number of detected nuclei, even when the training set includes 60\% detection noise (refer to Table \ref{Tab:ClassMetricsDetectionNoise}).
Negative and positive variations are observed for some metrics compared to the reference results obtained with V1.2, likely influenced by differences in the sets of nuclei detected.
The use of early training stopping, however, mitigates these variations, greatly reducing the loss of detected nuclei.

\begin{table*}[!ht]
    \centering
    \caption{Impact of different levels of detection noise introduced in the training set on the number of correctly detected nuclei in the test sets and on the balanced classification metrics computed on these nuclei, for models obtained with or without early training stop. The results are expressed as in Table \ref{Tab:DetectMetricsDetectionNoise}.}
    \begin{tabular}{@{}cc|c|cc|cc|cc@{}}
    \multicolumn{1}{c}{} &\multicolumn{1}{c}{} &\multicolumn{2}{c}{} \\ 
    \multicolumn{2}{c|}{\textbf{Classification}} & 
    \multicolumn{1}{c|}{V1.2 (clean)} &
    \multicolumn{2}{c|}{20\% detection noise}&
    \multicolumn{2}{c|}{40\% detection noise}&
    \multicolumn{2}{c}{60\% detection noise}\\ 
    \hline
    \multirow[c]{2}{*}{{Epochs}}
    & 1st stage &  25 & 14-17 & 25 & 5 & 25 & 1 & 25\\
    & 2nd stage &  50 & 4-6 & 50 & 3-4 & 50 & 1 & 50\\
    \hline
    \multirow[c]{1}{*}{{Accuracy}}
     &  Overall  & 96.8(0.3) & 95.0(0.3) & 92.8(0.6) & 94.8(0.5) & 92.7(0.7) & 95.2(0.6) & 89.6(1.1) \\
     \hline
    \multirow[c]{3}{*}{{Nuclei}}
     &  E  & 5101(144) & 5131(170) & 4553(117) & 4646(167) & 3000(85) & 4333(130.0) & 1660(61) \\
     &  L  & 5959(289) & 5956(283) & 5726(286) & 5642(272) & 4902(275) & 5292(263) & 2652(156)\\
     &  N  & 122(7) & 130(6) & 113(8) & 104(5) & 82(3) & 90(3) & 40(5)\\
    \hline
    \multirow[c]{3}{*}{{Precision}}
     &  E  & 96.2(0.7) & 97.0(0.7) & 94.7(1.2) & 98.9(0.1) & 96.4(0.8) & 97.6(0.5) & 99.0 (0.2) \\
     &  L  & 96.1(0.4) & 91.5(0.8) & 89.8(1.2) & 88.8(1.1) & 92.8(0.5) & 90.7(1.2) & 89.6(1.4) \\
     &  N  & 97.8(0.1) & 99.6(0.1) & 99.8(0.7) & 99.6(0.1) & 99.8(0.5) & 99.5(0.1) & 99.6(0.1) \\
    \hline
    \multirow[c]{3}{*}{{Recall}}
     &  E  & 97.6(0.4) & 94.3(0.8) & 96.9(0.4) & 96.1(0.6) & 94.7(0.6) & 95.5(0.7) & 91.5(0.9) \\
     &  L  & 96.9(0.4) & 98.7(0.1) & 96.3(0.5) & 98.2(0.2) & 94.4(0.8) & 98.7(0.3) & 96.3(0.3) \\
     &  N  & 95.9(0.5) & 92.0(0.5) & 85.3(1.2) & 90.0(0.1) & 89.0(1.4) & 91.3(1.2) & 81.1(3.1) \\
    \hline
    \multirow[c]{3}{*}{{F1-Score}}
     &  E  & 96.9(0.3) & 95.7(0.6) & 95.8(0.6) & 97.5(0.3) & 95.5(0.4) & 96.5(0.5) & 95.1(0.4) \\
     &  L  & 96.1(0.3) & 95.0(0.4) & 93.0(0.6) & 93.3(0.6) & 93.5(0.4) & 94.5(0.6) & 92.8(0.8) \\
     &  N  & 97.8(0.3) & 95.7(0.3) & 92.0(0.7) & 94.6(0.6) & 94.1(0.8) & 95.2(0.7) & 89.4(1.8) \\
     
    \end{tabular}

    \label{Tab:ClassMetricsDetectionNoise}
\end{table*}

In summary, detection noise, characterised by a substantial loss of training examples for each class, primarily influences detection performance, with segmentation performance being less affected. The robustness of classification performance is notable, likely attributed to the balanced and diverse data ensured by the weighted sampler and data augmentation during the constitution of training batches.
Across all analysed tasks, early training stopping proves to be a valuable strategy for mitigating the impact of detection noise. 

\subsection{Impact of segmentation noise on Hover-Net performance}\label{sec:segmentation_noise}
In this experiment, we introduce segmentation noise into the V1.2 training set by distorting the nuclei contours, and merging masks of adjacent nuclei of the same class. The latter modification results in the merging of 1-2\% of nuclei of each class.

As indicated in Table \ref{Tab:over_under_seg} in the Appendix, the introduction of this noise results in a decrease in over-segmentation occurrences and an increase in under-segmented occurrences in test set predictions, with or without early training stopping, albeit with a beneficial effect of the latter. Segmentation noise also amplifies the number of false-positive and false-negative detections, the former (FP) being exacerbated and the latter (FN) reduced by the use of early stopping.

These effects on detection lead to a deterioration in precision and recall compared to the model trained on a clean set, as shown in Table \ref{Tab:DetMetricsSegmentationNoise}. As observed earlier for detection noise (in Table \ref{Tab:DetectMetricsDetectionNoise}), the impact of early stopping on precision varies by cell type, but is beneficial on recall for all. These effects are more pronounced for neutrophils (likely due to their smaller numbers).

Concerning the segmentation metrics in Table \ref{Tab:SegMetricsSegmentationNoise}, calculated on the well-detected nuclei as detailed in Section \ref{sec:metrics}, there is a decline in performance compared to training with a clean set.
This decline is especially pronounced for lymphocytes and neutrophils, which, being small in size, experience a more pronounced impact from segmentation errors, particularly on metrics like IoU. Employing the early stopping strategy proves beneficial, notably enhancing performance in IoU. These variations in IoU closely correspond to the trends observed in under-segmentation errors, as illustrated in Table \ref{Tab:over_under_seg}.

Finally, segmentation noise has no major impact on classification performance, as detailed in the Appendix (see Table \ref{Tab:ClassMetricsSegmentationNoise}). 

\begin{table}[!ht]
    \centering
    \caption{Impact of segmentation noise (introduced in the training set) on the detection metrics computed on the test set. The results are expressed as in Table \ref{Tab:DetectMetricsDetectionNoise}.}
    \begin{tabular}{@{}c|c|c|cc@{}}
    \multicolumn{1}{c}{} &\multicolumn{1}{c}{} &\multicolumn{2}{c}{} \\ 
    \multicolumn{2}{c|}{\textbf{Detection}} & 
    \multicolumn{1}{c|}{V1.2 (clean)} &
    \multicolumn{2}{c}{Segment. noise} \\ 
    \hline
    \multirow[c]{2}{*}{{Epochs}}
    & 1st stage & 25 & 19-20 & 25\\
    & 2nd stage & 50 & 6-8 & 50\\
    \hline
    \multirow[c]{4}{*}{{Precision}}
    & E & 84.3(0.8) & 82.7(0.5) & 80.5(0.6)\\
    & L & 88.3(0.8) & 86.7(0.7) & 88.1(0.8)\\
    & N & 80.0(2.2) & 66.0(3.1) & 83.9(3.4)\\ 
    & Overall & 87.3(0.5) & 83.7(0.9) & 85.8(0.4)\\ 

    \hline
    \multirow[c]{4}{*}{{Recall}}
    & E & 85.9(0.7) & 80.3(2.1) & 78.7(0.8)\\
    & L & 92.9(0.3) & 90.0(0.8) & 89.5(0.2)\\
    & N & 85.9(0.4) & 88.3(2.0) & 81.8(2.5)\\ 
    & Overall & 89.5(2.3) & 85.4(1.3) & 84.2(0.5)\\ 

    \hline
    \multirow[c]{4}{*}{{F1-Score}}
    & E & 85.1(0.5) & 81.5(1.2) & 79.6(0.3)\\
    & L & 90.6(0.3) & 88.3(0.7) & 88.8(0.4)\\
    & N & 82.8(1.8) & 74.8(2.5) & 82.8(2.5)\\ 
    & Overall & 88.4(0.3) & 84.6(0.6) & 85.0(0.3)\\ 
    \end{tabular}
    \label{Tab:DetMetricsSegmentationNoise}
\end{table}

\begin{table}[!ht]
    \centering
    \caption{Impact of segmentation noise (in the training set) on the segmentation metrics computed on the test set. The results are expressed as in Table \ref{Tab:SegMetricsDetectionNoise}.}

    \begin{tabular}{@{}c|c|c|cc@{}}
    \multicolumn{1}{c}{} &\multicolumn{1}{c}{} &\multicolumn{2}{c}{} \\ 
    \multicolumn{2}{c|}{\textbf{Segmentation}} & 
    \multicolumn{1}{c}{V1.2 (clean)} &
    \multicolumn{2}{c}{Segment. noise}\\ 
    \hline
    \multirow[c]{2}{*}{{Epochs}}
     & 1st stage & 25 & 19-20 & 25\\
     & 2nd stage & 50 & 6-8 & 50 \\
    \hline
    \multirow[c]{4}{*}{{IoU}}
    & E & 80.8(0.3) & 76.9(1.0) & 76.7(0.1)\\
    & L & 76.6(0.2) & 72.6(0.1) & 69.8(0.2)\\
    & N & 76.5(0.6) & 74.6(0.6) & 67.8(0.9)\\ 
    & Overall & 78.0(0.2) & 74.7(0.6) & 71.4(0.3)\\
    \hline
    \multirow[c]{4}{*}{{HD}}
    & E & 4.7(0.1) & 5.7(0.2) & 5.7(0.1)\\
    & L & 3.0(0.0) & 3.6(0.1) & 3.9(0.1)\\
    & N & 4.3(0.0) & 4.6(0.2) & 4.9(0.2)\\ 
    & Overall & 3.7(0.1) & 4.5(0.1) & 4.7(0.1)\\

    \end{tabular}
    \label{Tab:SegMetricsSegmentationNoise}
\end{table}

\subsection{Impact of classification noise on Hover-Net performance}
We introduce varying levels of classification noise, low (20\%), medium (30\%), and high (40\%), into the training set. These percentages differ from those for detection noise, due to the severe impact that classification noise has on performance. 

Given the model's architecture, which combines outputs from three specialised branches of the decoder, classification can aid in segmenting adjacent nuclei when these nuclei correspond to different cell types. Classification noise, which can randomly affect the labels of clustered nuclei, can thus assist the network in correctly detecting them. Consequently, it is not surprising to observe a somewhat increased number of detected nuclei in the test sets with the introduction of classification noise in the training set (see Table \ref{Tab:ClassMetricsClassificationNoise}).
The data in Table \ref{Tab:over_under_seg} confirm a decrease in false-negatives, but with an increase in false-positives, and a beneficial impact of early stopping on both types of error. In contrast, the introduction of classification noise has no impact on the number of over- and under-segmented nuclei.

Table \ref{Tab:ClassMetricsClassificationNoise} illustrates the evolution of classification performance under different training conditions. 
Without early stopping, the influence of noisy class labels becomes noticeable as soon as a low level of classification noise is introduced into the training set. Remarkably, up to a medium noise level, the use of early training stopping almost restores the F1-scores observed in the absence of noise. The even greater negative impact of 40\% noise is also considerably reduced by the early stop of training.
As observed for detection noise, the number of epochs determined based on a clean validation set progressively decreases as the level of classification noise increases.

\begin{table*}[!ht]
    \centering
    \caption{Impact of different levels of classification noise (introduced in the training set) on the number of correctly detected nuclei and the balanced classification metrics (in the same way as in Table \ref{Tab:ClassMetricsDetectionNoise}).}
    \begin{tabular}{@{}cc|c|cc|cc|cc@{}}
    \multicolumn{1}{c}{} &\multicolumn{1}{c}{} &\multicolumn{6}{c}{} \\ 
    \multicolumn{2}{c|}{\textbf{Classification}} & 
    \multicolumn{1}{c|}{V1.2 (clean)} &
    \multicolumn{2}{c|}{20\% noise} &
    \multicolumn{2}{c|}{30\% noise}&
    \multicolumn{2}{c}{40\% noise}\\ 
    \hline
    \multirow[c]{2}{*}{{Epochs}}
     & 1st stage & 25 & 18-21 & 25 & 8-9 & 25 & 5-7 & 25\\
     & 2nd stage & 50 & 2-3 & 50 & 2 & 50 & 1-2 & 50\\
    \hline
    \multirow[c]{1}{*}{{Accuracy}}
     &  Overall & 96.8(0.3) & 96.1(0.4) & 90.2(0.5) & 96.4(0.2) & 83.9(0.6) & 92.1(0.4) & 69.6(1.2) \\
     \hline
    \multirow[c]{3}{*}{{Nuclei}}
     &  E  & 5101(144) & 5095(138) & 5052(143) & 5341(164) & 5213(143) & 5113(153) & 5138(146) \\
     &  L  & 5959(289) & 5979(186) & 5910(286) & 6032(291) & 6047(230) & 5989(287) & 5898(282)\\
     &  N  & 122(7) & 128(7) & 126(7) & 127(7) & 125(8) & 128(7) & 125(7)\\
    \hline
    \multirow[c]{3}{*}{{Precision}}
     &  E  & 96.2(0.7) & 98.6(0.2) & 90.8(0.7) & 97.8(0.4) & 89.8(1.5) & 99.5(0.1) & 67.2(2.5) \\
     &  L  & 96.1(0.4) & 92.6(1.0) & 87.9(0.9) & 94.3(0.4) & 77.0(0.7) & 84.9(0.8) & 72.9(0.7) \\
     &  N  & 97.8(0.1) & 97.6(0.2) & 94.9(0.7) & 97.5(0.2) & 89.8(0.5) & 94.3(1.3) & 69.6(1.1)  \\
    \hline
    \multirow[c]{3}{*}{{Recall}}
     &  E  & 97.6(0.4) & 94.7(0.8) & 92.5(0.3) & 95.5(0.3) & 85.9(0.5) & 87.6(1.3) & 66.6(1.1) \\
     &  L  & 96.9(0.4) & 97.6(0.2) & 96.1(0.3) & 97.6(0.2) & 93.0(0.4) & 99.2(0.1) & 77.3(1.6) \\
     &  N  & 95.9(0.5) & 95.9(0.4) & 82.1(1.5) & 95.9(0.4) & 72.8(1.6) & 89.5(0.9) & 64.7(3.7) \\
    \hline
    \multirow[c]{3}{*}{{F1-Score}}
     &  E  & 96.9(0.3) & 96.6(0.4) & 91.6(0.3) & 96.7(0.2) & 87.8(0.6) & 93.2(0.7) & 66.9(1.7) \\
     &  L  & 96.1(0.3) & 95.0(0.6) & 91.8(0.5) & 95.9(0.3) & 84.2(0.5) & 91.4(0.5) & 75.0(0.3) \\
     &  N  & 97.8(0.3) & 96.8(0.2) & 88.0(2.9) & 96.7(0.3) & 80.4(1.1) & 91.8(0.6) & 67.0(2.4) \\
     
    \end{tabular}
    
    \label{Tab:ClassMetricsClassificationNoise}
\end{table*}

\subsection{Impact of combining different types of annotation noise on Hover-Net performance}

In a first set of experiments, we introduce a medium level of detection noise (40\%) and classification noise (30\% on the remaining annotations) into the training set, i.e. the levels at which performance deterioration becomes more appreciable. 
This implies that out of the 27501 nuclei in the V1.2 training set, only 11551 remain unchanged, 11000 are removed, and the classes of 4950 are modified. 
Since classification noise primarily affects classification performance, we focus solely on classification metrics while specifying the number of correctly detected nuclei. 

In addition to the previously analysed scenario, involving pre-training the network with fairly similar and clean data (PanNuke), we also investigate the network's behaviour in the absence of such data, i.e., using a network pre-trained only with a general dataset (Imagenet).

\begin{table*}[!ht]
    \centering
    \caption{Impact of combining detection and classification noise on the number of correctly detected nuclei and the balanced classification metrics (in the same way as in Table \ref{Tab:ClassMetricsDetectionNoise}), and comparison of performance obtained using pre-training with PanNuke or with Imagenet alone.}
    \begin{tabular}{@{}cc|ccc|ccc@{}}
    \multicolumn{2}{c}{} &\multicolumn{3}{c}{PanNuke pre-training} &\multicolumn{3}{c}{Imagenet pre-training} \\ 
    \multicolumn{2}{c|}{\textbf{Classification}} & 
    \multicolumn{1}{c}{V1.2 (clean)} &
    \multicolumn{2}{c|}{40\% det 30\% class} &
    \multicolumn{1}{c}{V1.2 (clean)} &
    \multicolumn{2}{c}{40\% det 30\% class}\\ 
    \hline
    \multirow[c]{2}{*}{{Epochs}}
    & 1st stage & 25 & 3-5 & 25 & 50 & 35-37 & 50\\
    & 2nd stage & 50 & 2-5 & 50 & 50 & 3-6 & 50\\
    \hline
    \multirow[c]{1}{*}{{Accuracy}}
     &  Overall  & 96.8(0.3) & 95.9(0.2) & 70.5(0.5) & 96.2(0.4) & 92.2(0.7) & 71.8(1.0)\\
     \hline
    \multirow[c]{3}{*}{{Nuclei}}
     &  E  & 5101(144) & 5040(146) & 2844(102) & 5096(148) & 5089(160) & 2767(122) \\
     &  L  & 5959(289) & 5912(277) & 3982(173) & 5815(287) & 5751(261) & 4407(213)\\
     &  N  & 122(7) & 128(7) & 72(6) & 122(7) & 116(10) & 77(8) \\
    \hline
    \multirow[c]{3}{*}{{Precision}}
     &  E  & 96.2(0.7) & 98.7(0.1) & 80.2(0.4) & 97.1(0.4) & 99.0(0.4) & 90.0(1.9) \\
     &  L  & 96.1(0.4) & 90.7(0.4) & 79.7(1.0) & 93.1(0.9) & 87.3(2.6) & 69.6(1.8) \\
     &  N  & 97.8(0.1) & 99.1(0.1) & 85.3(0.7) & 99.8(0.1) & 92.5(0.1) & 90.0(0.9) \\
    \hline
    \multirow[c]{3}{*}{{Recall}}
     &  E  & 97.6(0.4) & 97.2(0.2) & 73.7(0.7) & 98.1(0.4) & 89.1(1.6) & 74.9(0.7) \\
     &  L  & 96.9(0.4) & 98.4(0.2) & 76.5(0.4) & 97.3(0.1) & 98.1(0.6) & 85.2(0.4) \\
     &  N  & 95.9(0.5) & 92.0(0.4) & 61.2(1.5) & 93.2(0.9) & 89.3(3.3) & 55.3(4.0) \\
    \hline
    \multirow[c]{3}{*}{{F1-Score}}
     &  E  & 96.9(0.3) & 97.9(0.2) & 76.8(0.4) & 97.6(0.2) & 93.8(1.0) & 81.7 (0.6) \\
     &  L  & 96.1(0.3) & 94.4(0.3) & 78.1(0.5) & 95.1(0.4) & 92.4(1.3) & 76.6(1.0) \\
     &  N  & 97.8(0.3) & 95.4(0.2) & 71.3(1.1) & 96.4(0.5) & 90.8(1.2) & 68.5(3.3) \\

    \end{tabular}
    \label{Tab:ClassMetricsCombiNoise}
\end{table*}

The results obtained using pre-trained models on PanNuke are reported in Table \ref{Tab:ClassMetricsCombiNoise} and Table \ref{Tab:over_under_seg}. These results emphasise that without early training stopping, the combination of detection and classification noise in the training set has a more significant impact on classification performance compared to the effects observed when each type of noise is applied separately. Again, this impact is considerably reduced when the training is stopped early. Regarding the numbers of detected nuclei, without early training stopping, they are close to those obtained in the presence of detection noise alone. With early training stopping, the model's detection ability improves, reaching levels comparable to those achieved by the model trained with the clean training set. 

In comparison, the models pre-trained with Imagenet (see Table \ref{Tab:ClassMetricsCombiNoise}) behave similarly, showing comparable noise effects that are advantageously reduced with early training stopping. Nevertheless, and as expected, in the presence of noisy annotations in the training set, initialising the network with weights trained on a similar and clean dataset slightly improves model performance and substantially reduces the number of training epochs determined on the basis of a clean validation set.

In a second set of experiments, we introduce all three types of annotation noise: 40\% of detection noise, 30\% of classification noise, and segmentation noise. To do this, we introduce segmentation noise (contour deformation and under-segmentation) to the remaining nuclei after introducing detection noise, and then modify the classes. 
Since the results in Section \ref{sec:segmentation_noise} indicate that segmentation noise has no particular impact on classification performance but does for the other tasks, we present the performance metrics for segmentation and detection in the Appendix, Section \ref{sec:combination_DetClassSeg}.

These results suggest that the combination of all three types of noise substantially degrades the training set, particularly for neutrophils, which constitute the minority class and are therefore the most affected in predictions, consistent with previous results. Early training stopping, however, still has a positive effect on performance.

Finally, we complement the above results with experiments targeting the PanNuke dataset after pre-training the model with the V1.2 MoNuSAC  dataset (i.e. reversing the roles played by the two datasets). The results are gathered in the Appendix (Section \ref{sec:pannuke_results}). Once again, the early training stopping strategy proves effective in mitigating the effects of annotation noise.

%% file: conclusions.tex
\section{Discussion}
\label{sec:discussion}

In this study, we address the issue of imperfect annotations in the context of multi-instance segmentation and classification in digital pathology applications. We investigate the influence of detection, segmentation, and classification errors on the performance of HoVer-Net by introducing varying levels of noise into the original annotations of a clean training set.
We chose to exclusively employ the HoVer-Net model because, as elaborated in Section \ref{section:Model}, it was developed specifically and remains a benchmark for this type of task. Moreover, a prior investigation in digital pathology \cite{foucart2020snow} indicates that imperfect annotations have a fairly comparable impact on the performance of various deep network architectures.

Using specific metrics to evaluate each task, our experiments reveal that each type of noise affects its corresponding task. Furthermore, detection noise has a noticeable impact on segmentation performance, while conversely segmentation noise affects detection performance. 
Interestingly, aside from being fairly robust to segmentation noise, classification performance also demonstrates resilience to detection noise, despite a marked reduction in the number of training samples available for each class. However, it is crucial to interpret this result in the context of the decreased number of nuclei correctly detected and subsequently classified in the test set. This becomes especially relevant as the combination of detection and classification noise amplifies the negative impact on classification performance. 

The influence of annotation noise intensifies when training is not appropriately terminated. The higher the level of noise in the annotations of the training set, the earlier the termination becomes necessary. For optimal effectiveness, such a training stop necessitates the use of a clean validation set, which can be quite relatively small. 
The second training stage, applied to the entire network, is particularly affected by annotation noise, reducing it to a few epochs. This stands in contrast to the noise-free scenario where this second stage takes the longest.

Additionally, we conducted a comparison between models pretrained on an uncorrupted set of similar data, PanNuke, and models pretrained on a general dataset, ImageNet. The results of this experiment revealed a significant reduction in the number of training epochs required for convergence, while avoiding overfitting to noise, when using PanNuke pretrained weights. This not only reduces computational costs during training but also enhances the model's predictive performance.
In our supplementary experiments targeting the PanNuke dataset (with pre-training on V1.2 MoNuSAC), we observed that the number and type of cells present in the training set have a notable impact on the model's performance and its robustness to annotation noise. These experiments further confirm that employing an early training stopping strategy based on a clean validation set contributes to improving the model's robustness.

Our findings emphasise the importance of investing resources in accurately annotating a small validation set before implementing complex strategies to manage noisy annotations. A well-annotated validation set proves effective in preventing overfitting to noisy annotations and contributes to maintaining model performance.
The literature suggests various alternative strategies, some of which face limitations in digital pathology. For instance, semi-supervised methods require accurate identification of correctly annotated data in the training set for the supervised stage.
In the context of multiple instance segmentation and classification, data filtering has been proposed for this purpose \cite{galvez2023cleaning}.
However, studies cited in Section \ref{section:review_approaches} indicate that semi-supervised approaches, even for simpler tasks like segmentation, offer no clear advantages \cite{foucart2020snow,jimenez2023computational}.
Alternatively, data initially excluded through filtering due to suspected annotation errors could be re-labelled by a model trained on data identified as correctly annotated. This approach, combining filtered and re-labelled data, appears promising for addressing missing annotations in simpler tasks such as segmentation \cite{foucart2020snow}.
Following our previous study on data filtering in the context of multiple instance segmentation and classification \cite{galvez2023cleaning}, we also explored this relabelling approach. Our (unpublished) findings indicate that using filtered data alone is just as effective.
The limited success of these alternative approaches may be explained by the advantageous impact of pre-training on similar data for CNNs like HoVer-Net, as demonstrated in our results. This pre-training ensures efficient feature extraction in the encoder part of the network, as evidenced by HoVer-Net's second training stage promptly stopping when the encoder is unfrozen in the presence of substantial annotation noise.
In the field of medical image analysis, transformers have recently been proposed to enhance feature extraction and semi-supervised approaches \cite{he2023transformers}, particularly in segmentation \cite{wang2023dealing,xiao2022efficient}. However, using these models in medical image analysis remains challenging due to resource-intensive requirements in terms of training data, computation, and memory \cite{he2023transformers}. Transformers also present specific challenges for histopathological image analysis \cite{Atabansi2023transformers}. Their contribution to the handling of imperfect annotations is seldom investigated and should be compared with that of CNNs, considering the ease with which the latter can leverage pre-training, early training stopping, and filtering of training data, as demonstrated in our study and elsewhere \cite{galvez2023cleaning}.

Despite conducting extensive experiments that offer valuable insights into the behaviour of a CNN under imperfect annotations during training, our work has certain limitations.  
Regarding classification noise, we acknowledge that our consideration of symmetric noise does not account for specific challenges an expert might face when annotating images. Consequently, a more comprehensive analysis should extend to comparing the impact of symmetric versus pair classification noise \cite{liang2022review}, incorporating more realistic errors.  
Future investigations should focus on enhancing the early training stopping strategy. Beyond representing various tissue sample origins, careful consideration for the validation set's selection and evaluation criterion could be implemented to balance the contribution of different instance classes. Given the major impact of detection and classification noise on performance, determining the optimal number of training epochs through metrics targeting instances to be detected and classified, in a class-balanced manner rather than their constituent pixels, would be worthwhile.  
Lastly, the observed positive effect of pre-training suggests that improving the encoder could enhance overall model performance. To test this hypothesis, exploring self-supervised methods with CNNs on one hand, or alternatively, with transformer-based architectures, particularly when large quantities of data are available, would be desirable.

\section{Conclusions}
\label{sec:conclusion}
This study investigates the robustness of deep neural networks to several sources of annotation noise in the complex task of multi-class instance segmentation and classification in medical images. It underscores the positive impact of having a small yet clean validation set on determining an appropriate number of training epochs, enhancing the model's resilience to annotation noise within the training set.
Additionally, pre-training on a clean dataset with similar characteristics further improves this robustness simultaneously reducing the required number of epochs. These findings complement those of the state-of-the-art literature mainly concerning the classification of small thumbnails \cite{song25does,bai2021understanding}.

\section{Declaration of competing interest}
The authors declare that they have no known competing financial interests or personal relationships that could have appeared to influence the work reported in this paper.

\section{Data availability}
All the data used comes from publicly available sources linked in the article.

\section{Acknowledgements}
LGJ acknowledges support from the Service Public de Wallonie Recherche under grant n° 2010235 - ARIAC by DIGITALWALLONIA4.AI. The CMMI is supported by the European Regional Development Fund and the Walloon Region (Wallonia- biomed; grant no. 411132-957270; project “CMMI-ULB”). CD is a senior research associate with the FNRS (Brussels, Belgium) and an active member of the TRAIL Institute of the Walloon-Brussels Federation.
The authors also thank Egor Zindy (DIAPath, ULB) for providing English writing support during the preparation of this manuscript.

%% file: appendix.tex
\clearpage
\newpage
\setcounter{table}{0}
\renewcommand{\thetable}{A.\arabic{table}}
\setcounter{figure}{0}
\renewcommand{\thefigure}{A.\arabic{figure}}
\appendix
\section{Appendix}
\subsection{Flowchart of the study}
\begin{figure*}
    \centering
    \includegraphics[width=\textwidth]{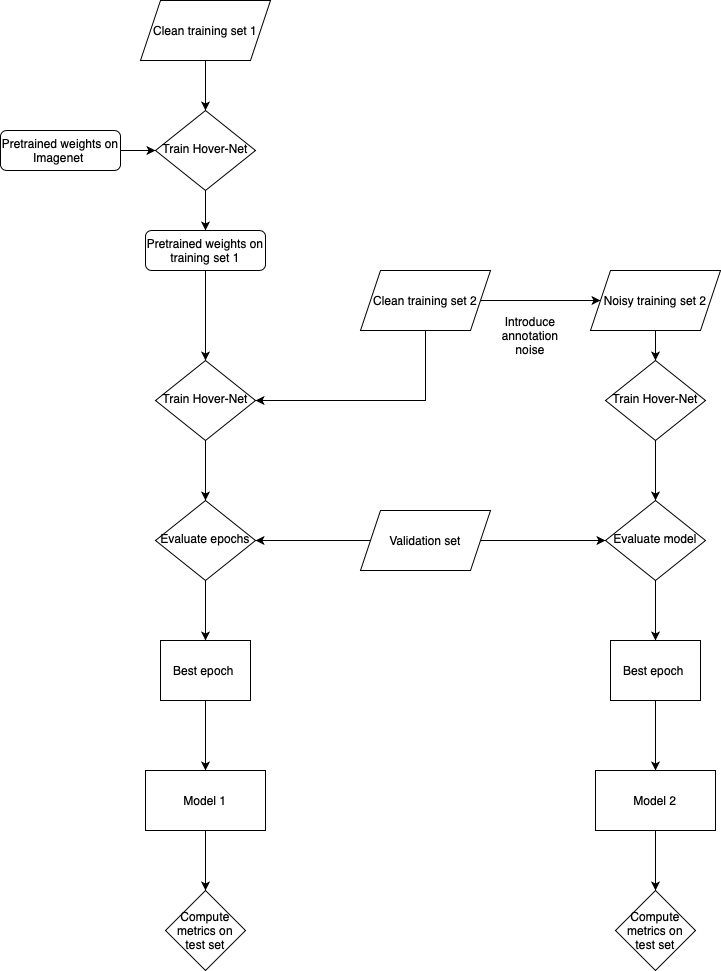}
    \caption{Flowchart of the experiments done (see details in the main text).}
    \label{flowchart}
\end{figure*}

See Figure \ref{flowchart}.

\subsection{Analysis of under- and over-segmentation}
See Table \ref{Tab:over_under_seg}

\subsection{Impact of segmentation noise on
classification performance}\label{sec:segmentation_DetClass}
See Table \ref{Tab:ClassMetricsSegmentationNoise}.

\subsection{Impact of combining detection, segmentation and classification noises on detection and segmentation performance}\label{sec:combination_DetClassSeg}
See Tables \ref{Tab:DetMetricsCombiNoise} and \ref{Tab:SegMetricsCombiNoise}.

\subsection{Experiments with the PanNuke dataset}\label{sec:pannuke_results}
We extend our investigation by replicating the experiments conducted with MoNuSAC, as detailed in the last results section, to the PanNuke dataset (refer to Section \ref{section:datasets}). Similarly, we introduce 40\% detection noise and 30\% classification noise into the training set, analyzing their impact on the number of correctly detected nuclei and classification performance. Due to the substantially larger quantity of annotated cells in the PanNuke dataset compared to MoNuSAC, and consequently a higher number of annotations unaffected by noise, we explore the effects of elevated levels of detection and classification noise—specifically, 60\% and 40\%, respectively.

In comparison with the MoNuSAC experiments, the results presented in Table \ref{Tab:ClassMetricsCombiNoisePannuke} confirm that detecting and classifying PanNuke nuclei without annotation noise is inherently more challenging. Furthermore, we observe that the introduced annotation noise in the PanNuke training set has a discernible impact, particularly at medium noise levels, which becomes more pronounced with higher noise levels.

Once again, our early training stopping strategy, based on a clean validation set, demonstrate effectiveness in mitigating the adverse effects of noise. It is worth noting that the complexity of the task for the PanNuke dataset slightly reduces the efficacy of this strategy. 

\begin{table*}[!ht]
    \centering
    \caption{Relationships under different training conditions between the number of detection errors, categorised as false positives (FP) and false negatives (FN), and the number of over- and under-segmentation cases in test set predictions. The results are expressed as in Table \ref{Tab:DetectMetricsDetectionNoise}.}
    \begin{subtable}{\textwidth}
    \centering
    \begin{tabular}{@{}cc|c|cc|cc|cc@{}}
    \multicolumn{1}{c}{} &\multicolumn{1}{c}{} &\multicolumn{2}{c}{} \\ 
    \multicolumn{2}{c|}{\textbf{Impact of a single noise type}} & 
    \multicolumn{1}{c|}{V1.2 (clean)} &
    \multicolumn{2}{c|}{40\% detection noise}&
    \multicolumn{2}{c|}{30\% classif. noise}&
    \multicolumn{2}{c}{Segment. noise}\\ 
    \hline
    \multirow[c]{2}{*}{{Epochs}}
    & 1st stage &  25 & 5 & 25 & 8-9 & 25 & 19-20 & 25\\
    & 2nd stage &  50 & 3-4 & 50 & 2 & 50 & 6-8 & 50\\
    \hline
    \multirow[c]{1}{*}{{FP}}
     &  & 1633(100) & 2019(88) & 1490(68) & 1884(100) & 3339(212) & 2076(166) & 1741(89)\\
    \multirow[c]{1}{*}{{Over-segmentation}}
     & & 78(6) & 78(2) & 81(4) & 79(7) & 76(6) & 25(1) & 28(3)\\
    \hline
    \multirow[c]{1}{*}{{FN}}
     & & 1309(35) & 2100(61) & 4508(116) & 992(24) & 1107(39) & 1821(171) & 1967(51)\\
     \multirow[c]{1}{*}{{Under-segmentation}}
     &  & 83(5) & 99(6) & 47(4) & 89(6) & 75(7) & 247(17) & 290(8)\\
    \end{tabular}
    \end{subtable}
    \begin{subtable}{\textwidth}
    \centering
    \begin{tabular}{@{}cc|cc|cc@{}}
    \multicolumn{1}{c}{} &\multicolumn{1}{c}{} &\multicolumn{2}{c}{} \\ 
    \multicolumn{2}{c|}{\textbf{Impact of combining several noise types}} & 
    \multicolumn{2}{c|}{40\% detect. 30\% classif.}&
    \multicolumn{2}{c}{40\% detect. 30\% classif. Segment.}\\ 
    \hline
    \multirow[c]{2}{*}{{Epochs}}
    & 1st stage & 3-5 & 25 & 7-9 & 25 \\
    & 2nd stage & 2-5 & 50 & 3-6 & 50 \\
    \hline
    \multirow[c]{1}{*}{{FP}}
     &  & 1422(71) & 1513(83) & 1549(203) & 1481(80)\\
    \multirow[c]{1}{*}{{Over-segmentation}}
     & & 48(7) & 82(7) & 51(7) & 75(8)\\
    \hline
    \multirow[c]{1}{*}{{FN}}
     & & 1412(29) & 5594(98) & 2062(232) & 5484(113)\\
     \multirow[c]{1}{*}{{Under-segmentation}}
     &  & 83(2) & 40(4) & 113(6) & 93(4)\\
    \end{tabular}
    \end{subtable}
    \label{Tab:over_under_seg}
\end{table*}

\begin{table}[!ht]
    \centering
    \caption{Impact of segmentation noise on the numbers of correctly detected nuclei and the corresponding balanced classification metrics (in the same way as in Table \ref{Tab:ClassMetricsDetectionNoise}).}
    \scalebox{0.95}{
    \begin{tabular}{@{}c|c|c|cc@{}}
    \multicolumn{1}{c}{} &\multicolumn{1}{c}{} &\multicolumn{2}{c}{} \\ 
    \multicolumn{2}{c|}{Classification} & 
    \multicolumn{1}{c|}{V1.2 (clean)} &
    \multicolumn{2}{c}{Segment. noise} \\ 
    \hline
    \multirow[c]{2}{*}{{Epochs}}
    & 1st stage & 25 & 19-20 & 25\\
    & 2nd stage & 50 & 6-8 & 50\\
    \hline
    \multirow[c]{1}{*}{{Accuracy}}
     &  Overall  & 96.8(0.3) & 96.2(0.3) & 95.5(0.5)\\
    \hline
    \multirow[c]{3}{*}{{Nuclei}}
     &  E  & 5101(144) & 4771(245) & 4671(145)\\
     &  L  & 5959(289) & 5774(250) & 5737(275)\\
     &  N  & 122(7) & 126(7) & 116(8)\\
    \hline
    \multirow[c]{3}{*}{{Precision}}
     &  E  & 96.2(0.7) & 98.3(0.3) & 96(0.4)\\
     &  L  & 96.1(0.4) & 93.6(0.6) & 91.8(1.2)\\
     &  N  & 97.8(0.1) & 99.6(0.1) & 99.8(0.1)\\
    \hline
    \multirow[c]{3}{*}{{Recall}}
     &  E  & 97.6(0.4) & 96.5(0.6) & 97.8(0.3)\\
     &  L  & 96.9(0.4) & 97.6(0.2) & 96.3(0.5)\\
     &  N  & 95.9(0.5) & 94.4(1.1) & 92.5(1.3)\\
    \hline
    \multirow[c]{3}{*}{{F1-Score}}
     &  E  & 96.9(0.3) & 97.4(0.4) & 96.8(0.3)\\
     &  L  & 96.1(0.3) & 95.6(0.2) & 94.0(0.6)\\
     &  N  & 97.8(0.3) & 96.9(0.6) & 96.0(0.7)\\
    \end{tabular}
    }
    \label{Tab:ClassMetricsSegmentationNoise}
\end{table}

\begin{table}[!ht]
    \centering
    \caption{Impact of combining detection, classification and segmentation noise on the detection metrics (in the same way as in Table \ref{Tab:DetectMetricsDetectionNoise}).}
    \scalebox{0.95}{
    \begin{tabular}{@{}c|c|c|cc@{}}
    \multicolumn{1}{c}{} &\multicolumn{1}{c}{} &\multicolumn{2}{c}{} \\ 
    \multicolumn{2}{c|}{\textbf{Detection}} & 
    \multicolumn{1}{c|}{V1.2 (clean)} &
    \multicolumn{2}{c}{40\% det 30\% class seg} \\ 
    \hline
    \multirow[c]{2}{*}{{Epochs}}
    & 1st stage & 25 & 7-9 & 25\\
    & 2nd stage & 50 & 3-6 & 50\\
    \hline
    \multirow[c]{4}{*}{{Precision}}
    & E & 84.3(0.8) & 86.0(1.2) & 84.2(0.9)\\
    & L & 88.3(0.8) & 88.4(2.0) & 84.1(0.9)\\
    & N & 80.0(2.2) & 16.5(3.8) & 15.6(1.8)\\ 
    & Overall & 87.3(0.5) & 87.1(1.1) & 82.3(0.4)\\ 

    \hline
    \multirow[c]{4}{*}{{Recall}}
    & E & 85.9(0.7) & 79.1(4.0) & 49.2(0.9)\\
    & L & 92.9(0.3) & 87.6(2.0) & 62.5(0.9)\\
    & N & 85.9(0.4) & 79.4(4.2) & 55.7(3.0)\\ 
    & Overall & 89.5(2.3) & 83.5(2.0) & 56.1(0.8)\\ 

    \hline
    \multirow[c]{4}{*}{{F1-Score}}
    & E & 85.1(0.5) & 82.3(1.2) & 62.1(0.7)\\
    & L & 90.6(0.3) & 88.0(0.1) & 71.7(0.6)\\
    & N & 82.8(1.8) & 27.0(4.9) & 24.4(2.4)\\ 
    & Overall & 88.4(0.3) & 85.2(0.8) & 66.8(0.5)\\ 
    \end{tabular}
    \label{Tab:DetMetricsCombiNoise}
    }
\end{table}

\begin{table}[!ht]
    \centering
    \caption{Impact of combining detection, classification and segmentation noise on the segmentation metrics (in the same way as in Table \ref{Tab:SegMetricsDetectionNoise}).}
    \scalebox{0.95}{
    \begin{tabular}{@{}c|c|c|cc@{}}
    \multicolumn{1}{c}{} &\multicolumn{1}{c}{} &\multicolumn{2}{c}{} \\ 
    \multicolumn{2}{c|}{\textbf{Segmentation}} & 
    \multicolumn{1}{c}{V1.2 (clean)} &
    \multicolumn{2}{c}{40\% det 30\% class seg}\\ 
    \hline
    \multirow[c]{2}{*}{{Epochs}}
     & 1st stage & 25 & 7-9 & 25\\
     & 2nd stage & 50 & 3-6 & 50 \\
    \hline
    \multirow[c]{4}{*}{{IoU}}
    & E & 80.8(0.3) & 76.5(1.1) & 74.6(0.5)\\
    & L & 76.6(0.2) & 72.0(0.7) & 71.3(0.2)\\
    & N & 76.5(0.6) & 70.8(3.2) & 71.7(1.1)\\ 
    & Overall & 78.0(0.2) & 73.1(1.1) & 72.6(0.2)\\
    \hline
    \multirow[c]{4}{*}{{HD}}
    & E & 4.7(0.1) & 5.5(0.2) & 7.1(0.2)\\
    & L & 3.0(0.0) & 3.5(0.1) & 3.8(0.0)\\
    & N & 4.3(0.0) & 5.1(0.5) & 4.9(0.4)\\ 
    & Overall & 3.7(0.1) & 4.5(0.1) & 4.9(0.1)\\

    \end{tabular}
    \label{Tab:SegMetricsCombiNoise}
    }
\end{table}

\begin{table*}[!ht]
    \centering
    \caption{Results for the PanNuke dataset (for models pretrained with V1.2 MoNuSAC): Impact of combinations of detection and classification noise on the number of correctly detected nuclei and the balanced classification metrics as in Table \ref{Tab:ClassMetricsCombiNoise}. }
    
    \begin{tabular}{@{}cc|c|cc|cc@{}}
    \multicolumn{1}{c}{} &\multicolumn{1}{c}{} &\multicolumn{2}{c}{} \\ 
    \multicolumn{2}{c|}{\textbf{Classification}} & 
    \multicolumn{1}{c|}{Clean} &
    \multicolumn{2}{c}{40\% det 30\% class}&
    \multicolumn{2}{c}{60\% det 40\% class}\\ 
    \hline
    \multirow[c]{2}{*}{{Epochs}}
    & 1st stage & 50 & 32-37 & 50 & 37-40 & 50 \\
    & 2nd stage & 50 & 27-29 & 50 & 13-15 & 50\\
     \hline
     \multirow[c]{1}{*}{{Accuracy}}
     &  Overall  & 78.8(0.1) & 74.4(0.1) & 72.9(0.2) & 66.7(0.1) & 62.2(0.1)\\
    \hline
    \multirow[c]{4}{*}{{Nuclei}}
     &  N  & 14381(320) & 13560(297) & 13042(301) & 12627(252) & 8968(195)\\
     &  I  & 7158(122) & 6917(108) & 6636(104) & 6420(118) & 5094(71)\\
     &  C  & 9859(169) & 9059(154) & 8746(146) & 8046(136) & 6045(89)\\
     &  E  & 5692(46) & 5454(55)  & 5226(56) & 4798(49) & 3882(39)\\
     \hline
    \multirow[c]{4}{*}{{Precision}}
     &  N  & 74.5(0.3) & 76.2(0.4) & 74.0(0.3) & 62.1(0.1) & 71.7(0.5)\\
     &  I  & 77.1(0.4) & 69.7(0.3) & 67.5(0.3) & 70.1(0.4) & 68.0(0.3)\\
     &  C  & 75.6(0.2) & 76.6(0.3) & 75.7(0.3) & 72.9(0.2) & 72.5(0.4)\\
     &  E  & 91.9(0.2) & 80.7(0.3) & 80.3(0.2) & 74.2(0.7) & 72.1(0.6)\\
    \hline
    \multirow[c]{4}{*}{{Recall}}
     &  N  & 89.3(0.1) & 79.2(0.4) & 80.1(0.3) & 76.2(0.4) & 63.9(0.4)\\
     &  I  & 80.4(0.2) & 83.7(0.1) & 83.8(0.2) & 76.7(0.2) & 75.7(0.3)\\
     &  C  & 67.0(0.3) & 58.5(0.2) & 56.6(0.4) & 54.1(0.3) & 50.0(0.2)\\
     &  E  & 78.6(0.5) & 76.3(0.5) & 71.1(0.7) & 59.8(0.5) & 59.2(0.4)\\
    \hline
    \multirow[c]{4}{*}{{F1-Score}}
     &  N  & 81.2(0.2) & 77.6(0.3) & 76.9(0.2) & 68.4(0.2) & 67.6(0.4)\\
     &  I  & 78.7(0.2) & 76.0(0.2) & 74.8(0.1) & 73.2(0.2) & 71.7(0.3)\\
     &  C  & 71.0(0.1) & 66.3(0.2) & 64.8(0.3) & 62.1(0.3) & 59.2(0.3)\\
     &  E  & 84.8(0.3) & 78.5(0.3) & 76.3(0.4) & 66.2(0.5) & 65.0(0.5)\\

    \end{tabular}
    \label{Tab:ClassMetricsCombiNoisePannuke}
\end{table*}